\documentclass[a4paper, journal, onecolumn, 12pt]{IEEEtran} 
\usepackage{amsmath,amsfonts}
\usepackage{array}
\usepackage{textcomp}
\usepackage{stfloats}
\usepackage{url}
\usepackage{verbatim}
\usepackage{graphicx}
\usepackage{cite}
\usepackage[breaklinks=true,colorlinks,citecolor=blue]{hyperref}
\usepackage{amssymb}
\usepackage{subcaption}
\usepackage{tabularx}
\usepackage{adjustbox}

\DeclareMathOperator*{\argmax}{\arg\max}   
\DeclareMathOperator*{\argmin}{\arg\min}

\usepackage{tikz}
\usepackage{tikz-layers}
\usepackage{tikz-3dplot}
\usepackage{tikzsymbols}
\tdplotsetmaincoords{60}{125}
\tdplotsetrotatedcoords{0}{0}{0} 
\usetikzlibrary{arrows.meta}
\usetikzlibrary{positioning, calc}
\usetikzlibrary{decorations.markings}
\tikzstyle arrowstyle=[scale=1.4]
\tikzstyle directed=[postaction={decorate,decoration={markings,
    mark=at position .65 with {\arrow[arrowstyle]{stealth}}}}]
\usepackage{forest}
\def\nodeminsize{8}
\def\nodedistance{30}
\tikzset{
    roundnode/.style={ellipse, draw=black!60, fill=black!2, thick, minimum size=\nodeminsize mm, node distance=\nodedistance mm, inner sep=1.75mm, font=\normalsize},
    rectanglenode/.style={rectangle, draw=black!60, fill=black!2, thick, minimum size=\nodeminsize mm, node distance=\nodedistance mm, inner sep=1.75mm, font=\normalsize},
    blanknode/.style={rectangle, draw=black!0, fill=black!0, thick, minimum size=\nodeminsize mm, node distance=\nodedistance mm, inner sep=1.75mm, font=\normalsize},
    edge from parent/.style={draw,->,black},
    line/.style={->},
    missionnode/.style={ellipse, draw=black!60, fill=black!5, thick, minimum size=8 mm, node distance=\nodedistance mm, inner sep=1.5mm, font=\normalsize, align=center}, 
}
\makeatletter
\newcommand{\gettikzxy}[3]{%
  \tikz@scan@one@point\pgfutil@firstofone#1\relax
  \edef#2{\the\pgf@x}%
  \edef#3{\the\pgf@y}%
}

\usepackage{algorithm, algorithmicx, algpseudocode}

\begin{document}

\title{Deployment of an Aerial Multi-agent System for Automated Task Execution in Large-scale Underground Mining Environments}

\author{Niklas Dahlquist, Samuel Nordström, Nikolaos Stathoulopoulos, Björn Lindqvist, Akshit Saradagi and George Nikolakopoulos
\thanks{\(^{1}\) Niklas Dahlquist is the corresponding author of the article {Email: \tt\small nikdah@ltu.se}. The authors are with Luleå University of Technology (LTU), Sweden. This work has been funded in part by the Swedish Department of Energy and LKAB through the Sustainable Underground Mining (SUM) Academy Programme project ”Autonomous Drones for Underground Mining Operations}%
}



\maketitle

\begin{abstract}
In this article, we present a framework for deploying an aerial multi-agent system in large-scale subterranean environments with minimal infrastructure for supporting multi-agent operations. The multi-agent objective is to optimally and reactively allocate and execute  inspection tasks in a mine, which are entered by a mine operator on-the-fly. The assignment of currently available tasks to the team of agents is accomplished through an auction-based system, where the agents bid for available tasks, which are used by a central auctioneer to optimally assigns tasks to agents. 
A mobile Wi-Fi mesh supports inter-agent communication and bi-directional communication between the agents and the task allocator, while the task execution is performed completely infrastructure-free. Given a task to be accomplished, a reliable and modular agent behavior is synthesized by generating behavior trees from a pool of agent capabilities, using a back-chaining approach. The auction system in the proposed framework is reactive and supports addition of new operator-specified tasks on-the-go, at any point through a user-friendly operator interface. The framework has been validated in a real underground mining environment using three aerial agents, with several inspection locations spread in an environment of almost 200 meters. The proposed framework can be utilized for missions involving rapid inspection, gas detection, distributed sensing and mapping etc. in a subterranean environment. The proposed framework and its field deployment contributes towards furthering reliable automation in large-scale subterranean environments to offload both routine and dangerous tasks from human operators to autonomous aerial robots.

\end{abstract}


\section{Introduction}\label{sec:introduction}

The use of autonomous robotic platforms 
in industrial production facilities is on the rise, both to increase profitability and to increase safety for human operators \cite{9916212}. Specifically, in deep underground mining, where the fundamental risk of accidents is high, the industry is focusing on creating a safer environment for humans 
by deploying robotic systems to either execute dangerous tasks or verify the safety before authorizing human entry.  
Through efforts in the mining industry, human workers have already been moved to safer locations in several critical operations via, for instance, teleoperation of heavy machinery. For tasks that still require direct human involvement, it is crucial to track the status of the environment to minimize the risk before executing the necessary work. Currently, many measures are taken to reduce risks to humans in mining environments, such as tags carried by the operators to keep track of all personnel inside the subterranean environment, fixed sensors to monitor seismic activity, air quality, etc. In this setting, autonomous aerial robotic inspection offers a revolutionary alternative, through which  
regular large-scale status monitoring and continuous data collection can be performed efficiently. 
The mining areas that require inspection are vast in size (anywhere from 100 m to several kilometers) and present a challenging scenario for aerial robotic inspectors with limited battery and traversal speeds. This necessitates the transition to deploying autonomous multi-agent robot systems for such large-scale operations~\cite{stathoulopoulos2024tfr}. 

The central problem in efficient and effective deployment of an aerial multi-robot system, in missions such as routine inspection, dangerous gas monitoring and change detection\cite{stathoulopoulos2023irregular}, is that of coordinating task execution between the participating robotic inspection agents. The coordinating entity/architecture is responsible for optimal distribution of available tasks 
among participating agents, and it is expected to be 
reactive, 
in order to handle sudden changes in both the environment and the overall mission plan. When considering the execution of the assigned task itself, a large portion of the development effort is spent on designing robust local behaviors associated with a particular mission. 
Therefore, it is valuable to utilize a modular design to allow for easy integration of new tasks or new robotic platforms into the coordinating architecture. This article presents an auction-based coordination architecture for aerial multi-agent systems to optimally distribute inspection tasks among available agents and a modular approach for synthesizing agent behaviors for task execution.   

Deploying multi-agent systems in subterranean mining environments usually implies that there exists limited communication possibilities, especially in certain areas, meaning that it is necessary to design the overall architecture to be resilient to degraded performance. In this work a big emphasis is put on i) designing the local autonomy of the individual agents, to make sure that tasks can be executed independently even in the case of communication failure, and ii) how to design the task allocation architecture, utilizing communication only for reactively allocating the available tasks, to enable large-scale missions in active underground mining environments. 

The performance of the proposed architecture has been validated by the deployment of a three-agent aerial robotic system in a large-scale mining environment to execute an inspection mission. Being able to perform full-system experiments in real operational environments is a great opportunity. We would like to thank our mining industry collaborators at Luossavaara-Kiirunavaara AB (LKAB), that operates the largest underground iron mine in the world. The field demonstrations included in this manuscript were enabled through the Sustainable Underground Mining (SUM) Academy programme, which is funded by LKAB and the Swedish Energy Agency, and were executed at a depth of over 500m in an active production area of the mine located in Kiruna, Sweden.





\begin{figure}[t]
    \centering
    \begin{subfigure}[t]{0.495\columnwidth}
        \centering
        \includegraphics[width = \textwidth]{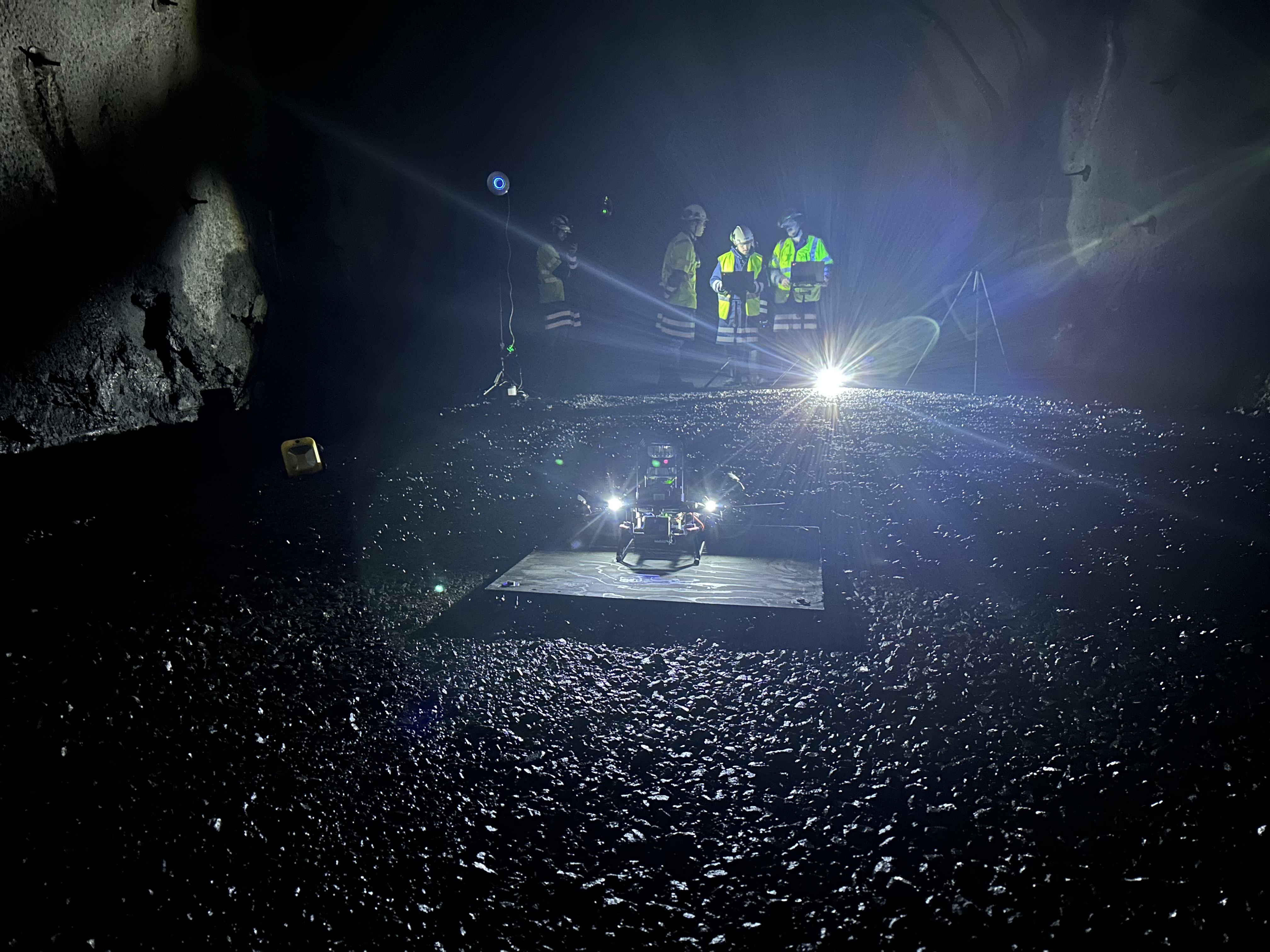}
        \caption{}
    \end{subfigure}%
    \hfill
    \begin{subfigure}[t]{0.495\columnwidth}
        \centering
        \includegraphics[width = \textwidth]{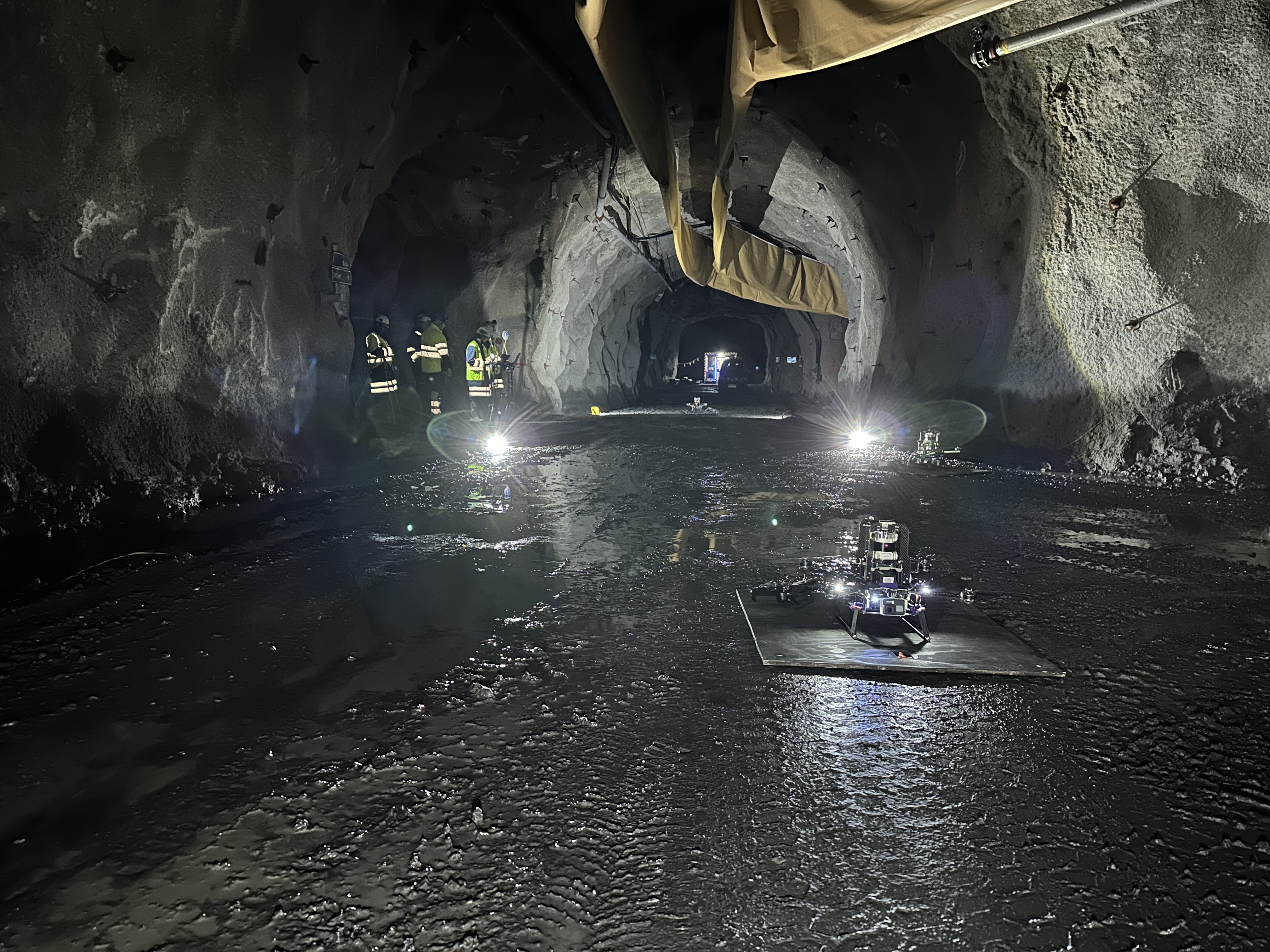}
        \caption{}
    \end{subfigure}
    \\
    \begin{subfigure}[t]{0.495\columnwidth}
        \centering
        \includegraphics[width = \textwidth]{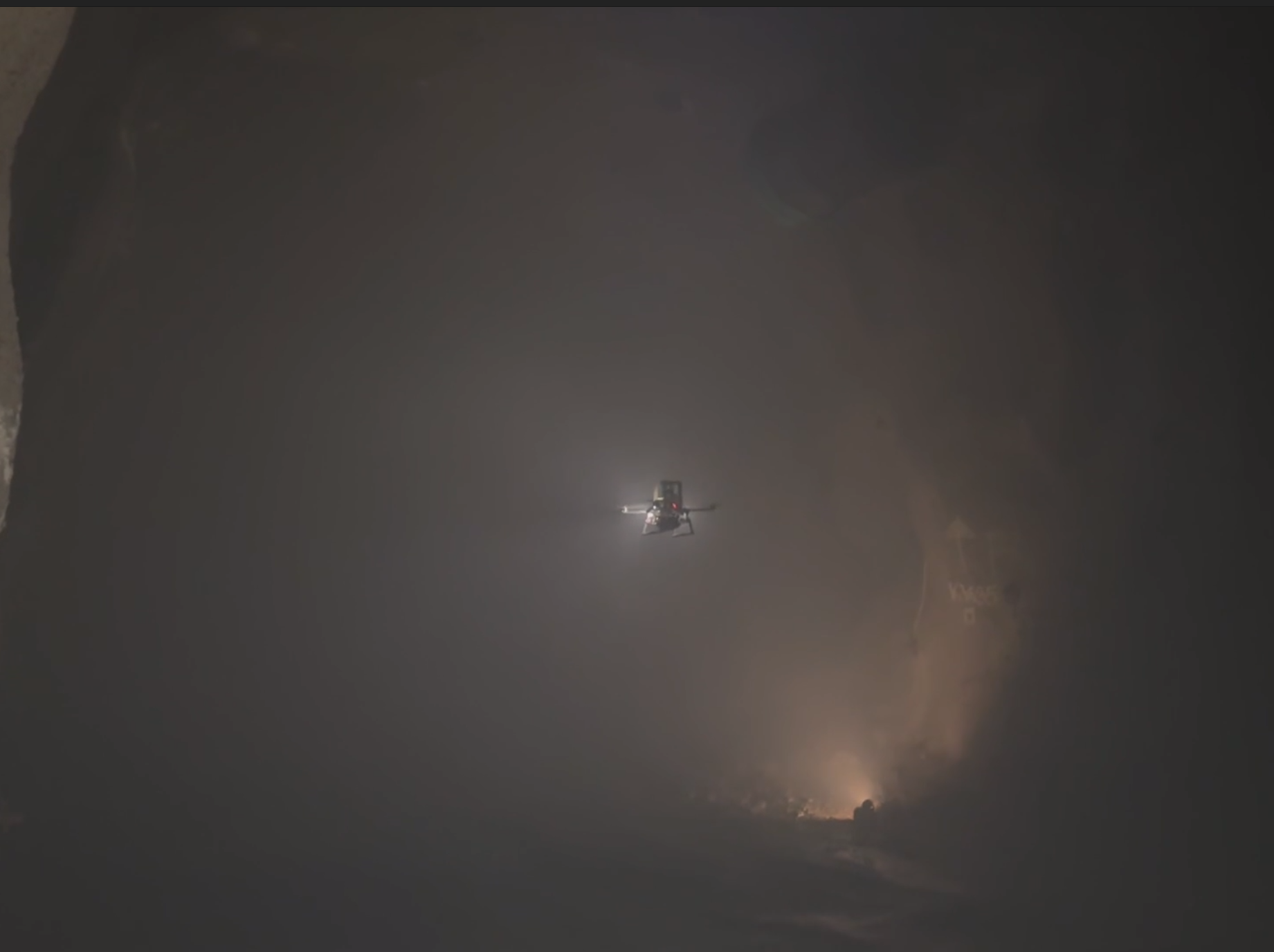}
        \caption{}
    \end{subfigure}%
    \hfill
    \begin{subfigure}[t]{0.495\columnwidth}
        \centering
        \includegraphics[width = \textwidth]{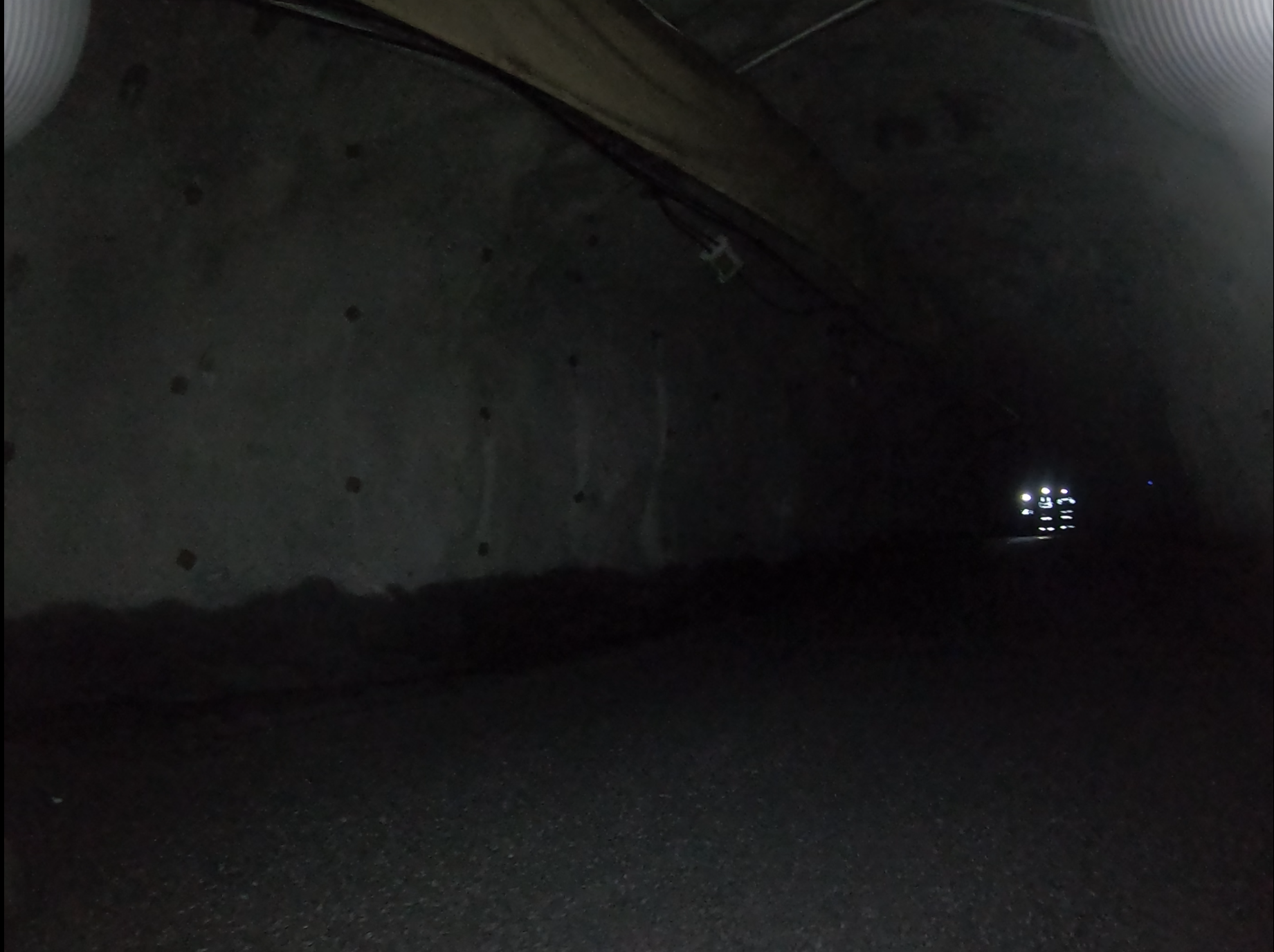}
        \caption{}
    \end{subfigure}%

    \caption{
    The evaluation environment at a depth of over 500m in an active production area of the mine located in Kiruna, Sweden. (a) and (b) show the tunnel environment, with the aerial vehicles on the ground before a mission is initiated, (c) illustrates one of the many challenges for autonomy in a mining environment, flying in smoke-filled, or dusty, environments. Finally, in (d), a snapshot of an empty tunnel is presented to show the scale of the operating environment.}
    \label{fig:subt_challenges}
\end{figure}

\subsection{Challenges for multi-agent deployment in underground mines}
Deploying robotic systems in underground mines, or in subterranean environments in general, comes with a specific set of challenges~\cite{Kamak2024darpa_slam} related to onboard perception and navigation systems summarized in Fig. \ref{fig:subt_challenges}. These types of environments are significantly perception-degraded; darkness limiting the use of visual sensors, and significant amounts of dust (or smoke in some use-cases) that can potentially corrupt LiDAR data. The environment geometry itself, consisting of narrow tunnels, significantly limits the field of view of onboard sensors, which, in combination with the self-similarity of the environment, poses significant challenges to the SLAM/localization system. Safe navigation and collision avoidance is also heavily stressed as the robot is constantly close to the tunnel walls or installed infrastructure, especially for aerial robots where any interaction with the environment leads to a crash - and sets definite requirements on the utilized navigation stack in terms of robot safety. 
The environment is highly dynamic, where heavy vehicles, ore piles, and infrastructure is constantly moved around to facilitate the extraction operations, while the tunnels themselves expand and change over time. As such, both the navigation and control stack as well as the overarching task coordination layer must be reactive to changes in the environment. For multi-agent coordination there are also limitations on the available connectivity and communication infrastructure. A completely centralized system for both task coordination and task execution would require real-time perception/mapping data from all agents in real time. In our proposed framework this challenge is addressed by limiting the communication requirements between agents and the coordinator to 1) available tasks (coordinator $\rightarrow$agents), 2) individual task costs (agents$\rightarrow$coordinator, 3) allocated tasks (coordinator$\rightarrow$agents), which is later visualized in Fig. \ref{fig:auction_summary}. 



\subsection{Related Works}
\subsubsection{Mining \& Subterranean Robotics}
While deploying aerial robots in underground  mines is difficult, the inherent safety risks and environmental challenges (vast size, 3D terrain etc.) also present significant opportunities for the technology. The following survey~\cite{shahmoradi2020comprehensive} discusses how aerial robots (at the time mainly teleoperated) have been used in underground mines for gas concentration measurements~\cite{dunnington2017fast}, visual and thermal inspection of rock mass characteristics~\cite{turner2018geotechnical}, inspection of the distribution of rock fragmentation (or muck piles) after blasting~\cite{bamford2017aerial}, and for providing visual inspection of unreachable areas~\cite{freire2017capture}. Towards the vision of fully automating such inspection tasks in real mining environments initial works exist, for example, in air quality monitoring or the detection of cracks in the tunnel walls, using autonomous ground vehicles~\cite{kim2022development, protopapadakis2016autonomous}. 
A large driving force in the recent developments and up-scaling of autonomous robotic missions in mining/subterranean environments can be attributed to the DARPA Subterranean Challenge~\cite{subt}, where teams deployed fleets of robots into fully unknown subterranean environments in the search-and-rescue context. During that period, the two central problems of subterranean localization/SLAM~\cite{palieri2020locus,chen2022direct,khattak2020complementary} and navigation~\cite{dang2019graph,palieri2020locus} were significantly extended. The challenge also saw the deployment of multi-robot systems~\cite{bayer2023autonomous} for collaborative mission execution. The robot collaboration was centered around multi-modal robots~\cite{lindqvist2022multimodality}, guiding robots not to explore overlapping areas through deployed communication nodes~\cite{rouvcek2020darpa}, multi-robot SLAM~\cite{chang2022lamp}, but most relevant to this manuscript also in the context of multi-robot task allocation~\cite{o2023dynamic}. 
In general, very few works exist in the academic literature that can showcase multi-robot collaborative deployment in the context of task allocation in real field mining environments. The framework presented in this manuscript not only deploys a novel task allocator and is demonstrated in a real field environment - it was also developed specifically to enable a non-expert user to easily interact with the system through an easy-to-use GUI which allows a single worker to operate the whole robotic fleet. 


The main difference between our approach and the one presented in \cite{o2023dynamic} stems from the different use-cases, our work is focused towards routine inspection and integrating aerial agents to be a part of the mining environment itself in contrast to focusing on exploring and searching for objects in a completely unknown environment. This sets our work apart since there exists some communication possibilities to enable a centralized operator to decide what tasks, in this work inspection of certain locations, need to be executed at what locations. This allows us to have a more predictable coordination strategy without needing to rely on a fully decentralized task allocation strategy.

Optimal multi-agent task allocation can be achieved through many approaches. The survey \cite{Rizk2019}, on multi-robot systems,   identifies four 
components in a typical complex multi-agent mission: 1) task decomposition, 2) coalition formation, 3) task allocation and 4) task execution/planning and control. This work focuses on designing a multi-agent coordination framework that focuses on the third and fourth components and assumes discrete tasks that do not require decomposition and can be accomplished by a single agent. 
In this work, we employ a market-inspired auction-based task allocation scheme to optimally assign available tasks to agents (for the reasons elaborated in section \ref{subsec:Task_Coordination}). The works   \cite{Dias-2004-8847, Das2015, Sariel-Talay2011} also utilize auctioning systems for multi-agent coordination, where the tasks are distributed among agents based on their bids. However, 
the validation of the auctioning systems is demonstrated through limited, laboratory-scale experiments. It is also worth noting that, none of the above cited works 
automate all four components of multi-agent coordination identified in \cite{Rizk2019}. Our  work in \cite{Auction-Behaviours_NDahlquist_CCTA_2023} presented a similar reactive auction-based task allocation strategy 
combined with behavior trees for tracking task execution, but without any field experiments for verifying the performance of the system and to ascertain the tranferability of the overall framework to a real-world large-scale scenario. In addition, a major difference between the current work and \cite{Auction-Behaviours_NDahlquist_CCTA_2023} is that in the current work, an aerial multi-agent system is deployed in a distributed way, with the individual agents having a full autonomy stack that is capable of executing tasks, by navigating in a large-scale mining environment. 
%
%
\subsubsection{Multi-Robot Task Execution Planning and Allocation}
To enable autonomous robots to execute complex tasks in diverse environments, it is essential to develop a flexible mechanism that utilizes the pool of modular robot behaviors to generate an optimized sequence of behaviors to accomplish the assigned task, while also considering the safety constraints imposed by the surroundings. Generating task plans involves the process of determining the sequence of actions and resources required to complete a task efficiently. Task plans can be generated dynamically based on current conditions and constraints, as in works such as \cite{9730031}, where task planning is accomplished considering only partial knowledge of the environment for the scenario where a robot needs to navigate and find specific objects. Classical methods such as Petri nets, decision trees, and state machines \cite{Holloway1997} have been widely utilized in task planning to model and formalize complex behaviors. These methods provide a structured approach to represent and analyze the behavior of automated systems. The main drawback of these methods is that, they lack modularity and a natural way to include reactivity. The authors of \cite{ghallab:hal-01982043} note that a big part of the currently available task planning methods results in a static plan of actions to solve the desired problem without the necessary flexibility to react to unforeseen events.

Behavior trees have recently gained a fair amount of success when in comes to deploying robotic systems in uncertain and dynamic environments due to their modularity and reactivity. Behavior trees are shown to generalize other common control structures, such as state machines and decision trees \cite{BT_introduction}, and require less programming efforts to develop and maintain due to its inherent modularity \cite{Iovino2023}. Recently, a method of generating reactive behavior trees for a robotic platform, based on a set of available actions and conditions, was presented in \cite{Colledanchise2019}. In this article, we adopt the theoretical method presented in \cite{Colledanchise2019} 
with several adjustments to the algorithm to enable generating a complete behavior tree prior to deployment, which were deemed necessary to verify and guarantee that the task planning would accomplish a given task. We present a large-scale experimental verification by deploying the method for aerial platforms which accomplish tasks in an underground mining environment, as part of a multi-agent system.
\subsection{Contributions}
In this work, we develop a flexible aerial multi-agent autonomy framework specifically suited for enabling task execution in underground mining environments. The contributions of this article in relation to available literature can now be stated as follows.
\begin{itemize}
    \item We present a reactive aerial multi-agent solution for simultaneous execution of routine inspection tasks 
    in an active underground mine, which is motivated by the industrial need for a safer and more efficient subterranean workspace. We experimentally demonstrate the merits of the proposed solution through large-scale field deployment in an underground mine which has no supporting infrastructure for multi-agent robotics. 
    \item The proposed methodology comprises of two main parts: i) global coordination orchestrated by a central auction-based task allocation system and ii) local autonomy centered around automated behavior tree generation, where a back-chaining approach is used to synthesize behavior trees from a pool of primitive agent capabilities. 
     We design a set of primitive capabilities required for autonomous aerial agents and utilize the back-chaining approach to enable inspection-style tasks. 
    \item Reliable integration of local autonomy and task allocation layers to enable reactive multi-agent coordination through: i) an easy operator interface, to allow for user-specific tasks to be added on-the-go, ii) deployment of a mobile Wi-Fi mesh to enable inter-agent communication and agent participation in the auctioning system, and iii) integration of the supporting elements of the autonomy stack, such as risk-aware path planning, global relocalization in a point-cloud and NMPC/APF-based tracking and collision avoidance.  
\end{itemize}
In addition, we include a discussion on structuring and executing large-scale multi-agent experiments in a mine, along with practical concerns and lessons learnt from our experience in deploying a team of aerial agents in an active underground mining environment in collaboration with the mining company Luossavaara-Kiirunavaara AB (LKAB). 

%

%
%

\section{Problem Overview}\label{sec:problem_overview}
In subterranean environments, deployment of an aerial multi-agent system on a large scale, with respect to both size of the environment and the number of agents, poses several challenges. Three of the most significant challenges that lead to the research problem addressed in this article are listed below.  

\textit{Reactive task allocation:} The foremost issues is ensuring efficient and optimal task allocation to the agents while being able to reactively re-distribute the available task amongst the agents, due to the uncertainty posed by a dynamic and unpredictable environment. Such a framework should be able to quickly adapt to a priori unknown factors, such as addition of new tasks, changes in the operating environment (unexpected obstacles), or agents failures during task execution.

\textit{Modular and flexible control architecture:} In addition to having a flexible task allocation framework, an efficient control architecture is essential to coordinate the actions of agents seamlessly to perform different kinds of available tasks. The need of having the capability to, on-the-fly, utilize the modular behaviors of individual agents to compose and plan the execution of complex tasks is apparent. 

\textit{Versatile and user-friendly operator interface:} To enable the deployment of such an autonomous multi-agent system in an industrial production environment, it is vital to have a versatile and easy-to-use operator interface to supervise and orchestrate the deployment. The operator interface could be used to introduce new tasks on-the-fly, command specific agents directly, set constraints based on the changing environment etc. Through the interface, the operators would possess valuable situational awareness / domain knowledge and will have the ability to interact and inform the task allocation process; Both when it comes to what actual tasks to perform, and the task execution order. In short, a single operator at the surface control center, potentially kilometers away, should easily be able to control a fleet of robotic agents through the user interface, that acts as the face of the proposed task assignment framework and robust mission-specific local autonomy.

In summary, the research problem that this article looks to tackle can be stated as follows. Design a reactive task allocation framework that is suitable for unpredictable underground environments, such as deep underground mines, that lends high level of autonomy to the agents while also allowing operators to interact and introduce tasks to the system on-the-go. To facilitate this, an efficient way of automatically structuring the agent autonomy has be designed, which at the agent level utilizes simple and modular agent capabilities to enable agents to perform many different types of tasks; and at the level of the multi-agent system enables easy integration of new tasks and capabilities into the system.
%
%
\section{Multi-agent Autonomy in SubT Environments}
This section presents three sub-components of the overall multi-agent autonomy and the expectations from the three sub-components when deploying in subterranean environments.
\subsection{Local Autonomy}
To deploy multi-agent systems in a constantly changing and uncertain environment, it is necessary to give a large degree of autonomy to the individual agents. In certain scenarios, the agents are designed to perform a discrete set of modular actions/behaviors, and this set of actions can vary from agent to agent in a heterogeneous multi-agent system.
A mechanism or methodology for optimally composing and sequencing the available modular actions into a control architecture for an agent to execute complex tasks is 
is crucial. Both the methodology and the resulting architecture are expected to have several features, including:
\begin{enumerate}
    \item \textit{Flexibility}: Different tasks may require different sets of actions or behaviors. By having a set of available actions for every agent, it is possible to mix the agents to create teams tailored to specific mission requirements. This flexibility allows for the adaptation of robotic systems to a wide range of tasks and environments.
    \item \textit{Reusability}: Developing new actions from scratch for each specific task can be time-consuming and resource intensive. A library of available actions allows developers to reuse and repurpose components that have already been created and tested. This can significantly reduce development time and costs to include new type of tasks.
    \item \textit{Modularity}: A library of actions encourages a modular approach to system design. Each action can be developed and tested independently, making it easier to debug and maintain the overall system. Additionally, modular architectures are inherently more scalable, allowing for easier inclusion of new functionality and sensing capabilities.
    \item \textit{Robustness}: By having a library of well-tested actions, developers can leverage existing implementations to create more robust control architectures. Actions that have been thoroughly tested and validated are less likely to contain bugs or unexpected behaviors, resulting in more reliable robotic systems.
\end{enumerate}

Behavior trees are well known for satisfying all the features mention above \cite{BT_introduction, Iovino2023}. The individual actions are collected as behaviors and are therefore inherently flexible, reusable and modular. The increased robustness stems from the ability to reuse certain well-tested pieces of code, even on multiple different platforms. In the context of the multi-agent subterranean mission scenario considered in this work and in the software architecture to be proposed in section \ref{sec:methodology}, behavior trees will be a central part of the local autonomy of the agents.


\subsection{Task Coordination} \label{subsec:Task_Coordination}

A reactive task allocation framework is necessary to handle uncertainty of the environment, specifically in the subterranean mining environment scenario considered in this work. For handling multi-agent systems such a framework is expected to be flexible to include different types of tasks, allow for heterogeneous agents to participate, dynamically adapt based on a changing environment, scale well with a large number of agents and give a considerable amount of autonomy to the agents to execute the tasks. The market-based task allocation architectures has been a classic approach \cite{Dias-2004-8847} for enabling the deployment of multi-agent systems in such environments. The features that  motivate the use of this multi-agent coordination approach include: 
\begin{enumerate}
    \item \textit{Flexibility}: The adaptable nature of auction-based task allocation enables real-time and on-the-fly adaptation to changing factors such as fluctuations in the set of available tasks, agent availability, and dynamically changing environments. This flexibility enables the system to quickly respond, and react, to new information and optimally reassign tasks to agents. This is especially important in the multi-agent mission considered in this scenario, where an operator can, at any point in time, add new tasks and the system needs to quickly adjust and optimally reallocate tasks.
    \item \textit{Fairness}: Auction-based methods provide a transparent and unbiased way of allocating tasks, ensuring that all participators have an equal chance to bid for, and get allocated to, the available tasks, based on their capabilities and current knowledge. The fairness enables trust between the agents and allows for agents with different capabilities to compete within the same system.
    \item \textit{Scalability}: Separation of optimal assignment (by the central auctioneer) and cost estimation (bids computed by the agents) into two separate processes makes an auction-based task allocation framework inherently scalable and capable of handling both large number of agents and large number of available tasks without suffering from performance degradation. This scalability makes it suitable for large scale deployment where reactivity is a priority, since it might be infeasible to plan the optimal task allocation over a long horizon.
    \item \textit{Decentralization}: By allowing agents to independently generate bids and affect the auction process through the bids, auction-based frameworks promote distributed decision-making. Having a partially decentralized decision process can lead to more robust and resilient systems that are less likely to fail due to single points of failure.
    \item \textit{Optimality}: Auction-based frameworks encourage agents to provide their best offers, leading to optimized task allocations that maximize overall system performance. Where reactivity is the main priority, auction based system can give an optimal solution over a short horizon, given the current information about the system and the environment.
\end{enumerate}
\subsection{Environment/Motivating Tasks}
Before describing the proposed framework in detail in the next section, we present a precise definition of the environment where the framework is expected to be deployed.

The overall multi-agent system is expected to be deployed in a deep underground mine, where the environment is both challenging and unique, with narrow and confined tunnels. The type of tasks that the multi-agent aerial system is expected to execute in the field deployment are ``inspection''  style tasks such as: gas monitoring, rock-face inspection etc. and the main target is to reactively allocate the tasks in a flexible and optimized manner. The choice of tasks executed are motivated by the need from the mining industry to optimize the rock-wall inspection when developing new tunnels, monitoring air quality after blasts for increased security and so on. Given this scenario, the algorithms and software modules developed in this work are tailor made for solving such an autonomous mission. The local autonomy of the agents is designed using automatically generated behavior trees due to the flexible and reactive nature of behavior trees and since it allows for rapid integration of additional tasks and robotic platforms without changing most of the developed software. A similar perspective guides the design of the auction-based task allocation, to maximize the autonomy given to the agents in the resulting framework. 
\subsubsection{Ease of Use for Operators}
For autonomous robotics to be accepted by the mining industrial community and to enable
frequent deployments in the field, 
an absolute requirement is to create a user-friendly interface for the operators, which allows them to supervise and interact with the framework, without the need to be trained in the details of the underlying framework.  
In this work, the task allocation framework is designed so that the only input that is required is the type of task to be executed and the location where the task has to be performed. This allows for an easy-to-use interface, where an operator can choose a location within a three dimensional map, where an inspection task is to be carried out. 
%
%
\section{Methodology}\label{sec:methodology}
\subsection{Automated Behavior Tree Synthesis for Agent Autonomy}
%
A behavior tree can be represented as a directed acyclic graph \(\mathrm{BT} = (\mathcal{N}, \mathcal{V})\), where \(\mathcal{N}\) is the set of nodes and \(\mathcal{V}\) defines the set of connections forming the behavior tree, examples of which can be found in Figures \ref{fig:Main_Behavior_Tree}, \ref{fig:bt_inspection_task} and \ref{fig:bt_default_behavior}. The leaf nodes are called \emph{execution nodes} and the internal nodes are called \emph{control nodes}. A behavior tree is executed by ticking the root node, this ``tick'' is passed down by the \emph{control nodes} until an \emph{execution node} receives it. A detailed description of the concept of behavior tree and the commonly used \emph{control nodes} can be found in \cite{BT_introduction}.

Consider the case where a library of available actions \(\mathcal{L} = \{\mathcal{A}_i, \mathcal{A^{\mathrm{Pre}}}_{i, j}, \mathcal{A^{\mathrm{Post}}}_{i, k}\}\) that a specific robotic platform can perform is known. The library \(\mathcal{L}\) consists of a set of actions \(\mathcal{A}_i\), where \(i \in \mathbb{N}\) is the number of available actions, a set of necessary conditions \(\mathcal{A^{\mathrm{Pre}}}_{i, j}\), where \(j \in \mathbb{N}\) represents the conditions that needs to be satisfied before action \(\mathcal{A}_i\) can be performed. The set of post conditions \(\mathcal{A^{\mathrm{Post}}}_{i, k}\), where \(k \in \mathbb{N}\) is the conditions that the action \(\mathcal{A}_i\) will satisfy after the actions is executed. It is assumed that there exists no cyclic dependencies between the pre conditions and the post conditions, meaning that we do not consider the case where an action would invalidate one or more of the conditions that are necessary to execute that said action. In addition, it is also assumed that the actions \(\mathcal{A}_i\) corresponds to behaviors that can be executed as \emph{execution nodes} and the pre- and post conditions can be implemented as \emph{condition nodes}.

\subsubsection{Behavior Tree Synthesis}

\begin{algorithm}
\caption{Pseudocode for generating the fully expanded behavior tree using the back-chaining approach.}
\label{alg:generate_tree}
\hspace*{\algorithmicindent} \textbf{Input:}  \(C_{\mathrm{start}}\) : Starting condition \\
\hspace*{\algorithmicindent} \textbf{Output:}  \(\mathrm{BT}\) : Final behavior tree
\begin{algorithmic}[1]
    \Function{GenerateBehaviorTree}{$C_{\mathrm{start}}$} 
        \State  $\mathrm{BT} \gets C_{\mathrm{start}}$ \Comment{Add initial condition}

        \While{$ConditionsToExpand \not\in \varnothing$}
            \State $C_{\mathrm{expand}} \gets getConditionToExpand(\mathrm{BT})$
            \State $\mathrm{BT} \gets Expand(\mathrm{BT}, C_\mathrm{expand})$
        \EndWhile
        \State \textbf{return} $\mathrm{BT}$ 
    \EndFunction
\end{algorithmic}
\end{algorithm}

\begin{algorithm}
\caption{Pseudocode for expanding a single condition.}
\label{alg:exapnd_condition}
\hspace*{\algorithmicindent} \textbf{Input:}  \(\mathrm{BT}\) : Behavior tree with condition to expand\\ 
\hspace*{\algorithmicindent} \hspace{1 cm} \(C_{\text{}}\) : Condition to expand \\
\hspace*{\algorithmicindent} \textbf{Output:}  \(\mathrm{BT}\) : The expanded behavior tree
\begin{algorithmic}[1]
    \Function{Expand}{$\mathrm{BT}, C$} 
        \State  \(\mathcal{A} \gets FindActionThatSolvesCondition(C)\)
        \State \([\mathcal{A^{\mathrm{Pre}}}_{1}, \dots , \mathcal{A^{\mathrm{Pre}}}_{n}] \gets GetPreConditionsToAction(\mathcal{A})\)
        \State \(T_\text{1} \gets Sequence([A^{\mathrm{Pre}}_{1}, \dots , A^{\mathrm{Pre}}_{n}], \mathcal{ A})\)
        \State \(T_\text{2} \gets Fallback(C, T_\text{1})\)
        \State \(\mathrm{BT} \gets Replace(\mathrm{BT}, C, T_\text{2})\)
        \State \textbf{return} $\mathrm{BT}$ 
    \EndFunction
\end{algorithmic}
\end{algorithm}

The principle of back-chaining, initially applied to behavior trees for the first time in \cite{Colledanchise2019}, stems from the idea that a goal conditions is specified and then, in a backward fashion, the necessary preconditions and actions are added. The algorithms \ref{alg:generate_tree} and \ref{alg:exapnd_condition} details how the algorithm is implemented, similar to \cite{Colledanchise2019}, but with the key difference that all conditions are expanded to make sure that it is feasible to achieve the desired condition before execution. In simple terms, the algorithms generates the task-specific behavior tree by expanding every condition by adding actions that satisfies that said condition until there are no more conditions to expand.

\subsubsection{Algorithm Steps in Detail}
i) Initialize the behavior tree (algorithm \ref{alg:generate_tree} line 2): The behavior tree is initialized by adding the final condition associated with completing the requested task. The behavior tree is then expanded until all actions required for satisfying this condition is added. 

ii) Run until all necessary conditions are expanded (algorithm \ref{alg:generate_tree} line 3 -- 6): This loop is executed until all conditions in the resulting behavior tree have been expanded according to its preconditions.

iii) Expand condition (algorithm \ref{alg:exapnd_condition}): This algorithm replaces a condition node \(C\) with a subtree including the necessary actions to complete accomplish said condition. It works as follows: First an action \(\mathcal{A}\) is selected that will satisfy the condition \(C\); second all precondtitions \([\mathcal{A^{\mathrm{Pre}}}_{1}, \dots , \mathcal{A^{\mathrm{Pre}}}_{n}] \) of \(\mathcal{A}\) are added in a sequence before the action \(\mathcal{A}\) and this subtree is called \(T_1\); Then, \(T_1\) is added in a fallback together with the original condition \(C\); Finally, the condition \(C\) is replaced in the behavior tree with the generated subtree \(T_2\).
\subsection{Auction-based Optimal Task Allocation}
In order to present the inner workings of the proposed task allocation architecture, this section begins by defining the necessary notations. In a multi-agent setting, the set of available tasks, at a specific time instant \(t\), is denoted by \( \mathfrak{T}_t = \{ T_1, T_2, \dots, T_{n_t} \} \), where \(n_t \in \mathbb{N}\) is the number of tasks available at time $t$. The tasks are considered to have one stage, for example ``inspect a location'', ``pick-up an object'' etc. A team of agents \(\mathfrak{R} = \{ R_1, R_2, \dots, R_{n_a} \}\) are available for executing the tasks, where \(n_a \in \mathbb{N}\) is the number of agents. We consider the setting where the tasks and their arrival times are not known a priori, i.e the set $\mathfrak{T}_t$ is a time-varying set, into which new tasks are added as they arrive and completed tasks are removed.
\begin{figure}[t]
    \centering
    \begin{subcaptionblock}[t]{0.33333\columnwidth}
        \centering
        \input{figures/tex/auction_fig_1}
        \caption{}
    \end{subcaptionblock}%
    \hfill
    \begin{subcaptionblock}[t]{0.33333\columnwidth}
        \centering
        \input{figures/tex/auction_fig_2}
        \caption{}
    \end{subcaptionblock}%
    \hfill
    \begin{subcaptionblock}[t]{0.33333\columnwidth}
        \centering
        \input{figures/tex/auction_fig_3}
        \caption{}
    \end{subcaptionblock}%
    \caption{An illustration of the different stages of the auction system. (a) Announcement stage. Available tasks \(\mathfrak{T}_t\) are announced to all agents. (b) Bidding stage. Agents calculate costs \(c_{k, j}\) and return them to the auctioneer. (c) Task allocation stage. The auction system optimizes how to assign tasks and broadcasts \(x_{k, j}\).}
    \label{fig:auction_summary}
\end{figure}

In this work, a market-inspired auction system, visually summarized in Fig. \ref{fig:auction_summary}, is employed for task allocation and it comprises of three stages: 
1) Announcement stage: The auctioning system announces the available tasks \(\mathfrak{T}_t\) to the agents. 
2) Bidding stage: The agents \(R_k\in\mathfrak{R}\) compute the costs (\(c_{k,j}\)) for accomplishing tasks $T_j\in\mathfrak{T}_t$ and send bids to the auctioning system.
3) Task allocation stage: The task allocator solves a combinatorial optimization problem to decide how the tasks \(\mathfrak{T}_t\) should be distributed among the agents \(\mathfrak{R}\). The allocated tasks are then announced to the agents.
Since the cost computations are performed locally by the agents, the central allocator has no need for the global information of the multi-agent system and this significantly reduced the complexity of the combinatorial task allocation problem and minimizes the communication requirements. Thanks to the modularity and simplicity of the auctioning system, the three constituent steps can be run at a high rate. This also keeps the communication overheads low by requiring the transmission of only: 1) the set of available tasks \(\mathfrak{T}\), 2) the bids from the agents \(c_{k, j}\) and 3) the allocated tasks \(x^*\) respectively during the three stages.
\subsubsection{Optimization Formulation}\label{subsec:Opti_Formulation}
To decide the optimal task allocation, the centralized task allocator solves a combinatorial optimization problem, specifically the following linear integer program (LIP):
\begin{equation}
    \begin{aligned}
         x^* = & \argmax_{x_{k, j}} \sum_{(k,j) \in E} \rho_{k,j} \cdot x_{k,j} \\
        \textrm{s.t.} \quad & \sum_{\{k \mid R_k \in \mathfrak{R}\}} x_{k,j} \leq 1, \text{ }  \forall j \in \{1, \ldots, n_t\} \\
        & \sum_{\{j \mid T_j \in \mathfrak{T}_t\}} x_{k,j} \leq 1, \text{ }  \forall k \in \{1, \ldots, n_a\}, \\
    \end{aligned}
    \label{eq:max_optimization_constrained}
\end{equation}
where \(x_{k, j}\in\{0,1\}\) is an assignment variable indicating if agent \(R_k\) is assigned to task \(T_j\) and the assignment vector \(x^\star\in\{0,1\}^{n_a \times n_t}\) defines the optimal distribution of tasks \(\mathfrak{T}_t\) among the agents \(\mathfrak{R}\). \(\rho_{k, j}\in\mathbb{R}\) represents the `profit' for agent \(R_k\) to complete task \(T_j\). $E$ is the set of two-tuples representing the edges of a bipartite graph between the set of agents and the set of tasks. The profits \(\rho_{k, j}\) are found by transforming the costs \(c_{k,j}\) in a way that reverses the order. This is done to make sure that the optimization problem assigns as many agents as possible to the available tasks. Here, it should be noted that the applicability of the optimization formulation \eqref{eq:max_optimization_constrained} can be easily expanded to more complex scenarios by including 
factors 
such as: relative importance of tasks, time prioritization of tasks and balancing the execution between different classes of tasks. This is done by scaling the additive terms appropriately or by adding additive terms and constraints corresponding to each individual factor. 

%
%
\subsubsection{Cost Estimation}
The estimation of the costs \(c_{k,j}\), for the available tasks \(\mathfrak{T}_t\) are performed locally by every agent \(R_k\). The cost \(c_{k,j}\) submitted by an agent, for completing the inspection tasks considered in this work, are based on the risk-aware path (elaborated in subsection \ref{subsec:Risk_Aware_Path_Planning}) from the current agent location to the inspection location. More specifically, the total distance of the risk-aware path from the current location of agent \(R_k\) to the inspection location of task \(T_j\) is considered as the total cost \(c_{k, j}\) to complete that task.
%
%
\subsection{Agent Autonomy} %
\begin{figure}
    \centering
        \begin{adjustbox}{max width = 0.8\columnwidth}
    \begin{tikzpicture}[
        align=center,
        ]

        \node[draw=black!60, fill=black!10, very thick, minimum width=0.9\columnwidth, minimum height = 19mm] (N1) at (0 mm, 0 mm) {N1};
        \node[draw=black!60, fill=black!10, very thick, anchor=north, minimum width=0.9\columnwidth, minimum height = 19mm] (N2) at ($(N1.south) + (0 mm, -5 mm)$) {};

        \node[anchor=north west] (N1_text) at (N1.north west) {Coordination Layer};

        \node[anchor=north west, draw=black!60, fill=green!10, very thick, minimum width = 0.25\columnwidth, minimum height = 6mm] (interface) at ($(N1_text.south west) + ( 0.0375 \columnwidth, 0 mm)$) {Operator\\interface};

        \node[anchor=north west, draw=black!60, fill=green!10, very thick, minimum width = 0.25\columnwidth, minimum height = 6mm] (task_assignment) at ($(interface.north east) + ( 0.0375 \columnwidth, 0 mm)$) {Task\\assigment};

        \node[anchor=north west, draw=black!60, fill=green!10, very thick, minimum width = 0.25\columnwidth, minimum height = 6mm] (visualization) at ($(task_assignment.north east) + ( 0.0375 \columnwidth, 0 mm)$) {Data\\visualization};


        \node[anchor=north west] (N2_text) at (N2.north west) {Local Autonomy (individual agents)};

        \node[anchor=north west, draw=black!60, fill=green!10, very thick, minimum width = 0.25\columnwidth, minimum height = 6mm] (auction_client) at ($(N2_text.south west) + ( 0.0375 \columnwidth, 0 mm)$) {Task\\execution};

        \node[anchor=north west, draw=black!60, fill=green!10, very thick, minimum width = 0.25\columnwidth, minimum height = 6mm] (path_planning) at ($(auction_client.north east) + ( 0.0375 \columnwidth, 0 mm)$) {Risk-aware\\path planning};

        \node[anchor=north west, draw=black!60, fill=green!10, very thick, minimum width = 0.25\columnwidth, minimum height = 6mm] (local_control) at ($(path_planning.north east) + ( 0.0375 \columnwidth, 0 mm)$) {Reactive\\navigation};

        \draw[line] ($ (N1.south) + (-0.4\columnwidth, 0)$) -- node[anchor=west] {\(\mathfrak{T}\), \(x^*\)} ($ (N2.north) + (-0.4\columnwidth, 0) $);
        \draw[line] ($ (N2.north) + (0.4\columnwidth, 0) $) -- node[anchor=west] {\(c_{k,j}\)} ($ (N1.south) + (0.4\columnwidth, 0)$);

    \end{tikzpicture}
    \end{adjustbox}

    
    \caption{ Local autonomy overview. Describes how the overall architecture operates. Highlighting the key components and that the coordination layer is responsible, based only on the estimated task execution costs \(c_{k, j}\), to allocate tasks \(x^*\) from the pool of available tasks \(\mathfrak{T}\). Fig. \ref{fig:fullSys} illustrates how every agent operates locally in greater detail.}
    \label{fig:autonomy_overview}
\end{figure}
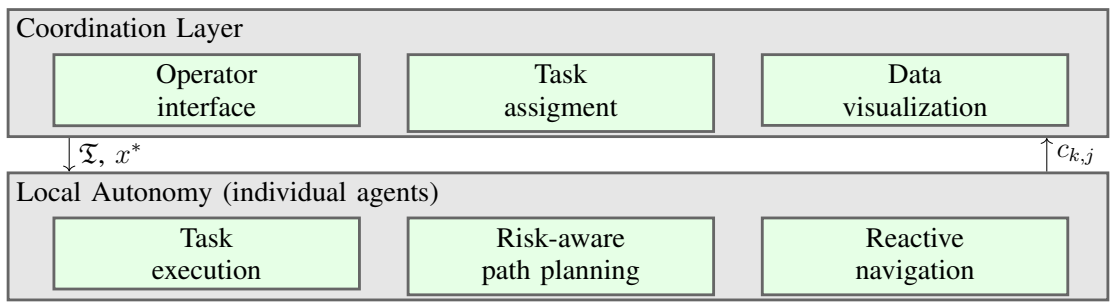

\begin{figure}
\centering
\begin{tikzpicture}[scale=1.25] 

    \coordinate (x1) at (-4.6, 0);
    \gettikzxy{(x1)}{\xa}{\oya};
    \coordinate (x2) at (-2.4, 0);
    \gettikzxy{(x2)}{\xb}{\oyb};
    \coordinate (x3) at (0, 0);
    \gettikzxy{(x3)}{\xc}{\oyc};
    \coordinate (x4) at (2.3, 0);
    \gettikzxy{(x4)}{\xd}{\oyd};
    \coordinate (x5) at (4.6, 0);
    \gettikzxy{(x5)}{\xe}{\oye};

    \coordinate (y1) at (7, 9);
    \gettikzxy{(y1)}{\oxa}{\ya};
    \coordinate (y1) at (7, 7);
    \gettikzxy{(y1)}{\oxab}{\yab};
    \coordinate (y2) at (6, 5.5);
    \gettikzxy{(y2)}{\oxb}{\yb};
    \coordinate (y3) at (4, 4);
    \gettikzxy{(y3)}{\oxc}{\yc};
    \coordinate (y4) at (2, 2);
    \gettikzxy{(y4)}{\oxd}{\yd};
    \coordinate (y5) at (0, 0);
    \gettikzxy{(y5)}{\oxe}{\ye};

    \draw[fill=yellow!20] (\xa - 11.5, \ya - 6.5) -- (\xa - 11.5, \ya + 6.5) -- (\xa + 10.2, \ya + 6.5) -- (\xa + 10.2, \ya + 3.2) -- (\xa + 13.5, \ya + 3.2) -- (\xa + 13.5, \ya - 6.5) -- cycle;
    \draw (\xa + 10.2, \ya + 6.5) -- (\xa + 13.5, \ya + 3.2);
    
    \draw[fill=yellow!20] (\xc - 46.5, \yab - 6.5+25) -- (\xc - 46.5, \yab + 6.5+25) -- (\xc - 24.8, \yab + 6.5+25) -- (\xc - 24.8, \yab + 3.2+25) -- (\xc - 21.5, \yab + 3.2+25) -- (\xc - 21.5, \yab - 6.5+25) -- cycle;
    \draw (\xc - 24.8, \yab + 6.5+25) -- (\xc - 21.5, \yab + 3.2+25);


    \node[] at (\xa, \ya) (bt) {.bt};
    \node[] at (\xc -35, \yab + 25) (pcd) {.pcd};

    \node[draw,fill=blue!10] at (\xc, \yab) (deg) {3DEG};
    
    \node[align=center,fill=green!10,draw] at (\xd, \ya) (lidar) {LiDAR\\IMU};
    
    \node[draw,fill=blue!10] at (\xc, \yb) (lio) {FAST-LIO};
    \node[draw, align=center,fill=blue!10] at (\xc, \yc) (vel) {Velocity\\[-1ex]Estimator};
    \node[draw,fill=blue!10] at (\xc, \ye) (nmpc) {NMPC};
    \node[draw,fill=blue!10] at (\xb, \ye) (apf) {APF};
    \node[draw, align=center] at (\xd, \ye) (px4) {PX4\\(low level\\control)};
    
    \node[draw,fill=blue!10] at (\xa, \yb) (octomap) {Octomap};
    \node[draw,fill=blue!10] at (\xa, \yd) (dsp) {D$^*_+$};
    \node[draw,align=center,fill=blue!10] at (\xb, \yc) (tsp) {Task execution\\[-1ex]behavior tree}; 
    \node[draw,fill=red!10] at (\xb, \ya ) (operator) {Auction System};
    \node[draw,fill=blue!10, align=center] at (\xa, \ye) (mux) {Mission\\[-1ex]executor};



    \draw[->] (bt) -- node[anchor=east] {$\mathbf{M}_{bt}$} (octomap);
    \draw[->] (octomap) -- node[anchor=east] {$\mathbf{M}_{bt}$} (dsp);
    \draw[->] (dsp) -- node[anchor=east, yshift=10] {$P$} (mux);
    \draw[->] (mux) -- node[anchor=south] {$P_{i}$}(apf);
    \draw[->] (nmpc) -- node[anchor=south, align=center] {} (px4); 
    \draw[->] (operator) -- node[anchor=east, align=center] {Assigned\\[-1ex]task} (tsp);
    \draw[->] (tsp) |- node[anchor=south east, yshift=20] {$WP_i$} (\xa+11,\yd+4); 
    \draw[->] (vel) |- node[anchor=south east, yshift=20] {$\hat{x}$} (\xa+11,\yd-4); 
    \draw[] (pcd) -- node[anchor=east, yshift=10] {$\mathbf{M}_{pcd}$} (\xc-35, \yb);
    \draw[->] (\xc-35, \yb) --  (lio);
    \draw[->] (\xc-35, \yab) --  (deg);
    \draw[->] (deg) -- node[anchor=west] {$T$} (lio);
    \draw[->] (\xd, \yab) -- node[anchor=south] {$\Pi_{pcl}$} (deg);
    \draw[->] (lio) -- node[anchor=east] {$\hat{p}, \hat{q}$}(vel);
    \draw[->] (vel) -- (nmpc);
    \draw[->] (lidar) |- node[anchor=east, yshift=10] {$\Pi_{pcl,IMU}$} (lio);
    \draw (lidar) -- node[anchor=east, yshift=-20] {$\Pi_{pcl}$} (\xd, 0.9);
    \draw[white, ultra thick] (\xc + 15, 0.9) -- (\xb + 10, 0.9);
    \draw[->] (\xd, 0.9) -| (apf);
    \draw[->] (apf) -- node[anchor=south] {$x_{ref}$} (nmpc);
    \draw[->] (\xc, 1.5) -| (\xa+8,\ye+11.5);
    \draw[->] (vel) -- node[anchor=south] {$\hat{x}$} (tsp);

\end{tikzpicture}

\caption{A block diagram describing the local autonomy that runs on every agent, to executes a task assigned by the auctioning system (shown in red). The blue boxes are the software that constitutes the autonomy stack. The sensors used by the agents are marked in green. The maps that are loaded from files are marked in yellow.
The white components are entities that take inputs from the agent-autonomy stack. PX4 is the onboard low-level controller that is directly controlling the aerial agents.}
\label{fig:fullSys}
\end{figure}
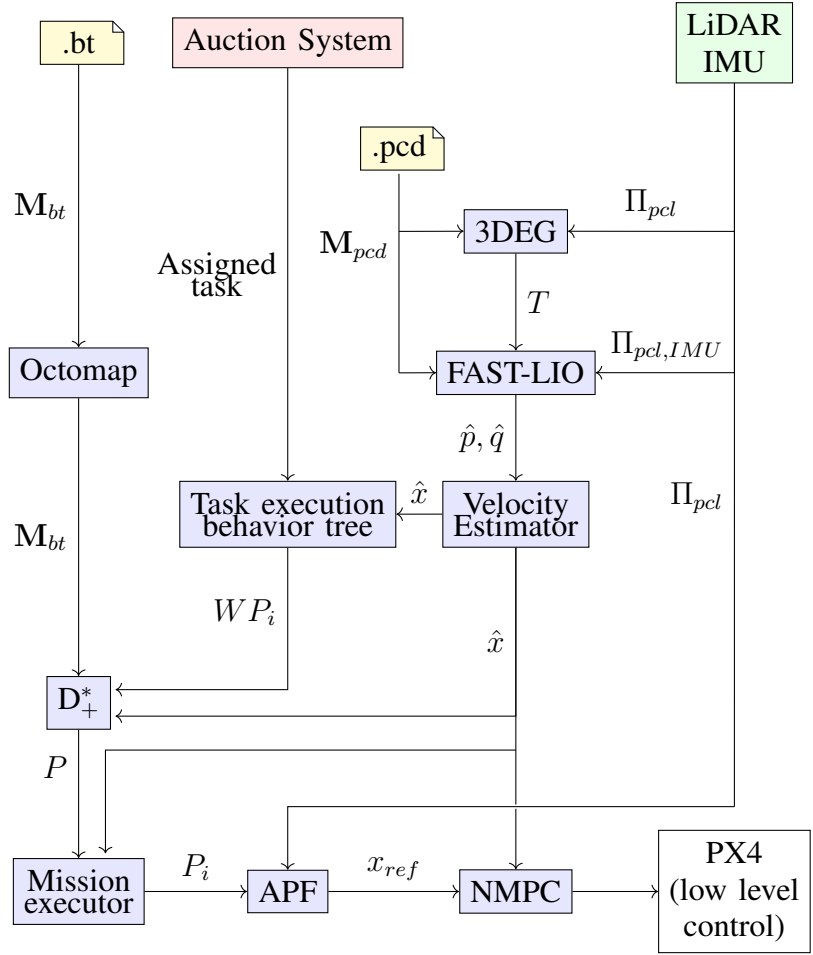

In this work, the local autonomy of the agents 
is designed to be able to operate independently, without relying on centralized infrastructure. Each agent manages its own operation and reacts to the environment to ensure resilient and robust operation. This approach allows the task execution to be handled in a completely infrastructure-free environment, only utilizing communication infrastructure to enable communication with the central auctioneers responsible for optimal task allocation. Fig. \ref{fig:autonomy_overview} shows a high level description of how the local autonomy communicates with the coordination layer and Fig. \ref{fig:fullSys} details the modules of the local autonomy running on every agent.

%
%
\subsubsection{Global and Local Localization}~\label{subsec:localization}

To localize the agents, we utilize the publicly available LiDAR-Inertia Odometry framework, FAST-LIO~\cite{xu2021fast}, which provides odometry at 10Hz by fusing LiDAR and IMU data. It is optimized for computational efficiency and robustness, making it ideal for UAVs with limited payload capacity. FAST-LIO uses a tightly-coupled iterated extended Kalman filter to integrate LiDAR feature points with IMU data, facilitating navigation in high-speed, noisy, and cluttered environments. The method for calculating the Kalman gain is designed to be efficient, as it depends on the state dimension rather than the measurement dimension, thereby lowering the computational burden. 
In a multi-agent system, effective and efficient task execution as well as waypoint navigation require localization within a common coordinate frame. To align the local odometry from FAST-LIO with the world coordinate frame $\mathcal{W}$, an initialization process transforms the local robot frame $\mathcal{R}$ to the world frame $\mathcal{W}$ through a homogeneous rigid transformation of the special Euclidean group, $T: \mathcal{R} \rightarrow \mathcal{W}$. The extension of FAST-LIO\footnote{\url{https://github.com/HViktorTsoi/FAST_LIO_LOCALIZATION}} requires an initial guess $T_0$ to perform a two-step ICP (Iterative Closest Point) registration and computation of the refined transformation $T$ before starting the odometry estimation. Operators can manually provide the initial guess for each agent, especially if missions begin from unique starting points. Alternatively, a fixed starting position can be set to a predetermined initial location, though this introduces unnecessary constraints and reduces the system's autonomy.

To address this challenge, frameworks similar to our previous work, 3DEG~\cite{stathoulopoulos2023deg} can be utilized. Given a point cloud map, 3DEG can estimate the transformation matrix $T \in SE(3)$ that aligns the current robot frame $\mathcal{R}$ with the global map frame $\mathcal{W}$. Using a point cloud map $M_{pcd} \in \mathbb{R}^3$ and its corresponding trajectory $Tr \in \mathbb{R}^3$, 3DEG determines the rigid transformation matrix $T$ from a single LiDAR scan. The process starts by partitioning the map and deriving the set of $n$ descriptive vectors $Q = \{\vec q_1, \vec q_2, \ldots, \vec q_n\}$, where each element $\vec q_i \in \mathbb{R}^{64}$ and $W = \{\vec w_1, \vec w_2, \ldots, \vec w_n\}$, where each element $\vec w_i \in \mathbb{R}^{64}\}$. The vectors $\vec q_i$  encode location-specific information, while the vectors $\vec w_i$ encode orientation-specific data, enabling the estimation of the yaw difference between point cloud scans.
Next, a sample from the LiDAR is used to extract the current descriptor, $\vec q_t$. This descriptor is then used to query the map's vectors, identifying the nearest neighbor $\vec q^*_t$ through the following optimization routine:
\begin{equation} \label{eq:argmin}
\vec q^*_t = \argmin_{\vec q_i \in Q}f(\vec q_i, \vec q_t),
\end{equation}
where the cost function $f: \mathbb{R}^{64} \times \mathbb{R}^{64} \rightarrow \mathbb{R}$ captures ...  
The index $i$ of the nearest neighbor $\vec q^*_t$ in the set $Q$ is used to find the corresponding vector $\vec w^*_i$ from $W$, which are together used to construct the initial guess transform $T_0$, which is then passed to FAST-LIO. Finally, FAST-LIO uses this estimate to perform the two-step ICP algorithm, refining the alignment between the robot frame and the current map, ensuring precise localization.


\subsubsection{Risk-aware Path Planning} \label{subsec:Risk_Aware_Path_Planning}
\begin{figure}
    \centering

\begin{tikzpicture}[scale=0.7,tdplot_rotated_coords,
                    cube/.style={very thick,black},
                    grid/.style={very thin,gray},
                    axis/.style={->,blue,ultra thick},
                    rotated axis/.style={->,purple,ultra thick}]

aw[cube,fill=green!5] (0,0,0) -- (3,0,0) -- (3,0,3) -- (0,0,3) -- cycle;
    

\foreach \x in {0,1,2,3,4,5,6,7}
   \foreach \y in {0,1,2,3,4,5,6,7,8,9}
      \foreach \z in {0,1,2}{
           \ifthenelse{  \lengthtest{\x pt < 7pt}  }
           {
                \draw [black]   (\x,\y,\z) -- (\x+1,\y,\z);
           }
           {
           }
           \ifthenelse{  \lengthtest{\y pt < 9pt}  }
           {
                \draw [black]   (\x,\y,\z) -- (\x,\y+1,\z);
           }
           {
           }
           \ifthenelse{  \lengthtest{\z pt < 2pt}  }
           {
                \draw [black]   (\x,\y,\z) -- (\x,\y,\z+1);
           }
           {
           }
}    
    
    \shade[ball color=red!70!white] (0.5,0.5,1.5) circle (.2);

     \draw[ultra thick, directed] (2.5, 0.5, 1.5) -- (0.5, 0.5, 1.5);
     
 \definecolor{high}{rgb}{1,0.3,0.3}
 \definecolor{mid}{rgb}{0.7,0.7,0.3}
     \definecolor{low}{rgb}{0.3,1.0,0.3}
    \draw[cube,fill=darkgray, opacity=0.9, draw=none] (0,4,0) -- (1,4,0) -- (1,4,2) -- (0,4,2) -- cycle;
    \draw[cube,fill=darkgray, opacity=0.9, draw=none] (1,3,0) -- (1,4,0) -- (1,4,2) -- (1,3,2) -- cycle;
    
    \draw[cube,fill=high, opacity=0.6, draw=none] (0,5,2) -- (0,5,0) -- (2,5,0) -- (2,5,2) -- cycle;
    \draw[cube,fill=high, opacity=0.6, draw=none] (2,2,0) -- (2,5,0) -- (2,5,2) -- (2,2,2) -- cycle;
    
    \draw[cube,fill=low, opacity=0.6, draw=none] (0,1,2) -- (0,6,2) -- (3,6,2) -- (3,1,2) -- cycle;
    \draw[cube,fill=high, opacity=0.6, draw=none] (0,2,2) -- (0,5,2) -- (2,5,2) -- (2,2,2) -- cycle;
    \draw[cube,fill=darkgray, opacity=0.9, draw=none] (0,3,2) -- (0,4,2) -- (1,4,2) -- (1,3,2) -- cycle;
    
    \draw[cube,fill=low, opacity=0.6, draw=none] (0,6,2) -- (0,6,0) -- (3,6,0) -- (3,6,2) -- cycle;
    \draw[cube,fill=low, opacity=0.6, draw=none] (3,1,0) -- (3,6,0) -- (3,6,2) -- (3,1,2) -- cycle;

    \draw[ultra thick, directed] (3.5, 7.5, 1.5) -- (2.5, 6.5, 1.5);
     \draw[ultra thick, directed] (2.5, 6.5, 1.5) -- (2.5, 2.5, 1.5);
     \draw[ultra thick, directed] (2.5, 2.5, 1.5) -- (3.5, 1.5, 1.5);
     \draw[ultra thick, directed] (3.5, 1.5, 1.5) -- (2.5, 0.5, 1.5);
    
    \draw[cube,fill=darkgray, opacity=0.9, draw=none] (6,5,1) -- (6,6,1) -- (5,6,1) -- (5,5,1) -- cycle;
    \draw[cube,fill=darkgray, opacity=0.9, draw=none] (6,5,1) -- (5,5,1) -- (5,5,2) -- (6,5,2) -- cycle;
    \draw[cube,fill=darkgray, opacity=0.9, draw=none] (6,6,0) -- (5,6,0) -- (5,6,1) -- (6,6,1) -- cycle;
    \draw[cube,fill=darkgray, opacity=0.9, draw=none] (7,6,0) -- (6,6,0) -- (6,6,2) -- (7,6,2) -- cycle;
    \draw[cube,fill=darkgray, opacity=0.9, draw=none] (7,0,2) -- (7,6,2) -- (6,6,2) -- (6,0,2) -- cycle;
    \draw[cube,fill=darkgray, opacity=0.9, draw=none] (7,0,0) -- (7,6,0) -- (7,6,2) -- (7,0,2) -- cycle;
    \draw[cube,fill=darkgray, opacity=0.9, draw=none] (6,4,2) -- (6,5,2) -- (5,5,2) -- (5,4,2) -- cycle;
    
    \draw[cube,fill=high, opacity=0.6, draw=none] (6,0,2) -- (6,4,2) -- (5,4,2) -- (5,0,2) -- cycle;
    \draw[cube,fill=high, opacity=0.6, draw=none] (5,3,2) -- (5,6,2) -- (4,6,2) -- (4,3,2) -- cycle;
    \draw[cube,fill=high, opacity=0.6, draw=none] (7,6,2) -- (7,7,2) -- (5,7,2) -- (5,6,2) -- cycle;
    \draw[cube,fill=high, opacity=0.6, draw=none] (6,5,2) -- (6,6,2) -- (5,6,2) -- (5,5,2) -- cycle;
    \draw[cube,fill=high, opacity=0.6, draw=none] (5,6,1) -- (5,7,1) -- (4,7,1) -- (4,6,1) -- cycle;
    
    \draw[cube,fill=high, opacity=0.6, draw=none] (5,6,0) -- (4,6,0) -- (4,6,2) -- (5,6,2) -- cycle;
    \draw[cube,fill=high, opacity=0.6, draw=none] (7,7,0) -- (5,7,0) -- (5,7,2) -- (7,7,2) -- cycle;
    \draw[cube,fill=high, opacity=0.6, draw=none] (5,7,0) -- (4,7,0) -- (4,7,1) -- (5,7,1) -- cycle;
    \draw[cube,fill=high, opacity=0.6, draw=none] (7,6,0) -- (7,7,0) -- (7,7,2) -- (7,6,2) -- cycle;

    \draw[cube,fill=low, opacity=0.6, draw=none] (5,0,2) -- (5,3,2) -- (4,3,2) -- (4,0,2) -- cycle;
    \draw[cube,fill=low, opacity=0.6, draw=none] (4,2,2) -- (4,7,2) -- (3,7,2) -- (3,2,2) -- cycle;
    \draw[cube,fill=low, opacity=0.6, draw=none] (7,7,2) -- (7,8,2) -- (4,8,2) -- (4,7,2) -- cycle;
    \draw[cube,fill=low, opacity=0.6, draw=none] (5,6,2) -- (5,7,2) -- (4,7,2) -- (4,6,2) -- cycle;
    \draw[cube,fill=low, opacity=0.6, draw=none] (4,7,1) -- (4,8,1) -- (3,8,1) -- (3,7,1) -- cycle;
    \draw[cube,fill=low, opacity=0.6, draw=none] (4,8,0) -- (3,8,0) -- (3,8,1) -- (4,8,1) -- cycle; 
    
    \draw[ultra thick, directed] (4.5, 8.5, 0.5) -- (3.5, 7.5, 1.5);
    
    \draw[cube,fill=low, opacity=0.6 , draw=none] (4,7,1) -- (3,7,1) -- (3,7,2) -- (4,7,2) -- cycle;
    \draw[cube,fill=low, opacity=0.6, draw=none] (7,8,0) -- (4,8,0) -- (4,8,2) -- (7,8,2) -- cycle;

    \draw[cube,fill=low, opacity=0.6, draw=none] (7,7,0) -- (7,8,0) -- (7,8,2) -- (7,7,2) -- cycle;
    
    \draw [black]   (0,1,2) -- (3,1,2);
    \draw [black]   (3,1,2) -- (3,6,2);
    \draw [black]   (3,6,2) -- (0,6,2);
    \draw [black]   (0,1,2) -- (0,6,2);
    \draw [black]   (0,2,2) -- (2,2,2);
    \draw [black]   (2,2,2) -- (2,5,2);
    \draw [black]   (2,5,2) -- (0,5,2);
    
    \draw [black]   (0,3,2) -- (1,3,2);
    \draw [black]   (1,3,2) -- (1,4,2);
    \draw [black]   (1,4,2) -- (0,4,2);
    
    \draw [black]   (0,6,2) -- (0,6,0);
    \draw [black]   (3,6,0) -- (0,6,0);

    \draw[black] (4,0,2) -- (4,2,2);
    \draw[black] (4,2,2) -- (3,2,2);
    \draw[black] (3,2,2) -- (3,7,2);
    \draw[black] (3,7,2) -- (4,7,2);
    \draw[black] (4,7,2) -- (4,8,2);
    \draw[black] (4,8,2) -- (7,8,2);
    
    \draw[black] (3,8,1) -- (3,7.3,1);
    \draw[black] (4,8,1) -- (3,8,1);
    \draw[black] (4,8,2) -- (4,8,1);
    \draw[black] (3,7,2) -- (3,7,1.3);
    
    \draw[black] (4,0,2) -- (7,0,2);
    \draw[black] (7,0,2) -- (7,8,2);
    \draw[black] (5,0,2) -- (5,3,2);
    \draw[black] (5,3,2) -- (4,3,2);
    \draw[black] (4,3,2) -- (4,6,2);
    \draw[black] (4,6,2) -- (5,6,2);
    \draw[black] (5,6,2) -- (5,7,2);
    \draw[black] (5,7,2) -- (7,7,2);
    
    \draw[black] (6,0,2) -- (6,4,2);
    \draw[black] (6,4,2) -- (5,4,2);
    \draw[black] (5,4,2) -- (5,5,2);
    \draw[black] (5,5,2) -- (6,5,2);
    \draw[black] (6,5,2) -- (6,6,2);
    \draw[black] (6,6,2) -- (7,6,2);
    
    \draw[black] (7,0,2) -- (7,0,0);
    \draw[black] (7,6,2) -- (7,6,0);
    \draw[black] (7,7,2) -- (7,7,0);
    \draw[black] (7,8,2) -- (7,8,0);
    \draw[black] (3,8,1) -- (3,8,0);
    
    \draw[black] (3,8,0) -- (7,8,0);
    \draw[black] (7,8,0) -- (7,0,0);
    
    \draw[black] (1,7,2) -- (3,7,2);
    \draw[black] (1,6,2) -- (1,7,2);
    \draw[black] (2,6,2) -- (2,7,2);
    \draw[black] (2,8,2) -- (2,7,2);
    \draw[black] (3,8,2) -- (3,7,2);
    \draw[black] (2,6.3,1) -- (2,7,1);
    \draw[black] (1,6,1) -- (1,7,1);
    \draw[black] (2,7,2) -- (2,7,1);
    
    \draw[black] (5,9,2) -- (7,9,2);
    \draw[black] (5,8,2) -- (5,9,2);
    \draw[black] (6,8,2) -- (6,9,2);
    \draw[black] (7,8,2) -- (7,9,2);
    \draw[black] (4,8,1) -- (4,9,1);
    \draw[black] (5,8,1) -- (5,9,1);
    \draw[black] (6,8,1) -- (6,9,1);
    \draw[black] (7,8,1) -- (7,9,1);
    \draw[black] (5,9,2) -- (5,9,1);
    \draw[black] (6,9,2) -- (6,9,1);
    \draw[black] (7,9,2) -- (7,9,1);

    
\shade[ball color=blue!70!white] (6.5,8.5,0.5) circle (.2);
\draw[ultra thick, directed] (6.5, 8.5, 0.5) -- (4.5, 8.5, 0.5);
\end{tikzpicture}

        \caption{An illustration of how D$^*_+$ plans a path from the blue ball to the red ball, by taking the risk layer into account. The red voxels are assigned higher risk, the green voxels are assigned lower risk, and black voxels are occupied.}
        \label{fig:riskViz}
    \end{figure}
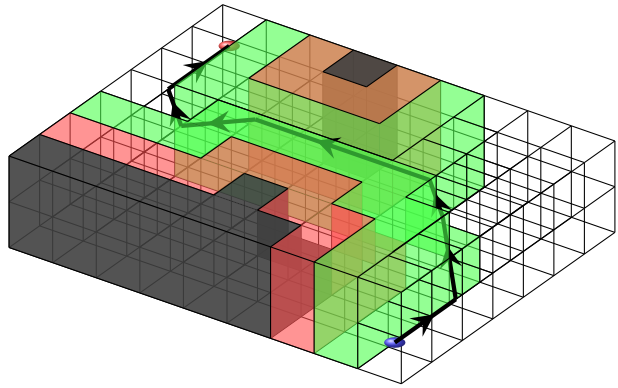

To plan a path from the current location of an agent to the location when an assigned task has to be executed, 
the path planner D$^*_+$ \cite{karlsson2023d} is used. D$^*_+$ is a risk-aware path planner that considers unknown areas and areas close to obstacles and walls, as a risk. For such areas, in computing the traversal cost $c$, additional costs are assigned to account for risks, over the standard distance costs/heuristics. D$^*_+$ is based on the occupancy voxel map provided by OctoMap~\cite{hornung2013octomap}, and the resulting path plan $P$ is a list of voxels that a UAV must sequentially visit, to move from point `A' to point `B'. D$^*_+$ plans the cheapest path $P$ between two points in the map, by minimizing the sum of the traversal cost along a path $P$, i.e, $\sum_{c_i \in P} c_i $. The traversal cost $c$ for each voxel is calculated as follows:
\begin{align}
    c_r &= \begin{cases}
        \frac{c_u}{d + 1} & \text{if $d < r$} \\
        0 & \text{else} \\
    \end{cases}\\
    c &= \begin{cases}
        c_r + c_d& \text{if the voxel is free} \\
        c_u + c_r + c_d& \text{if the voxel status is unknown}
    \end{cases}
    \label{eq:c}
\end{align}
where $d$ is the distance of the voxel from the closest occupied voxel and $r$ is the radius in which risk is evaluated. $c_d$ is the standard cost that is proportional to the travel distance or length of the path. The cost for unknown voxels $c_u$ and the risk cost $c_r$ are part of the total cost to traverse through that voxel, which is denoted as $c$. The path plan $P$ provided by D$^*_+$, is the shortest and safest path, as all the three costs $c_r$, $c_d$ and $c_u$ are considered in planning a path. 
An illustrative case of D$^*_+$ planning is shown in Fig.~\ref{fig:riskViz}. Result from the deployment of D$^*_+$ in a subterranean field experiment is presented in Fig.~\ref{fig:pathComp}, where both 
the planned path and the path tracked by an aerial robot are illustrated, showing that D$^*_+$ planning chooses the safest route whenever possible. 

In this article, the agents estimate the cost for executing an available task by computing the length of the D$^*_+$ path $P$ that safely takes a UAV agent 
from its current location `A' to the target location `B' associated with an available task. The cost computed by the agents are sent as bids 
to the central auction-based task allocation system for incorporating in the optimization problem \eqref{eq:max_optimization_constrained} in section \ref{subsec:Opti_Formulation}. 
D$^*_+$ is chosen because of its capability to reliably produce safe-to-navigate paths for aerial agents. 
\subsubsection{Reactive Path Tracking \& Local Avoidance} \label{subsec:NMPC_APF}
\begin{figure}[htb!]
    \centering
    \includegraphics[width=0.95\columnwidth]{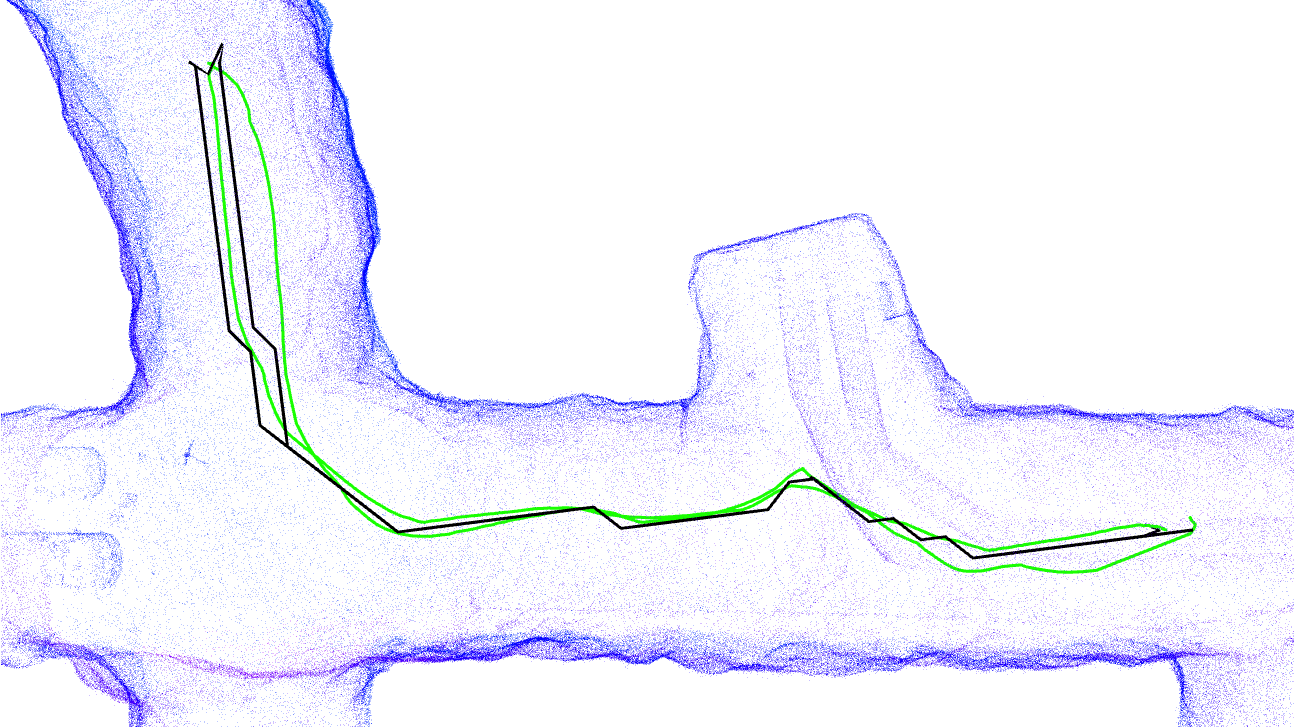}
    \caption{A comparison between the planned path (black line) and the path the aerial agent followed (green line) during navigation. Using the local avoidance and path tracking, $P$ is followed closely with only minor deviation while staying at a safe distance from any obstacle.}
    \label{fig:pathComp}
\end{figure}

Each individual agent uses a reactive local navigation kit~\cite{lindqvist2022adaptive} to track D$^*_+$ paths to target inspection locations. The kit is composed of a high-performance nonlinear model predictive controller (NMPC) combined with a fully reactive Artificial Potential Field (APF). The APF uses raw 3D LiDAR scans to generate repulsive forces (avoidance maneuvers) from obstacles and walls. The kit acts as an additional safety layer to ensure robot safety while traversing the tunnels to the inspection locations, resulting in small insignificant deviation from the planned path as can be seen in Fig.~\ref{fig:pathComp}.
\subsubsection{Behavior Tree Structure}
\begin{figure}
    \centering
    \begin{adjustbox}{max width = \columnwidth}
    \begin{forest}
    for tree={align=center, edge={-{Latex[length=1.5mm]}, draw}, parent anchor=south,child anchor=north, l=50pt}
    [\large \(\rightarrow\), rectanglenode
        [\large ?, rectanglenode
            [Current inspection completed?, roundnode, style={draw, fill=cyan!10}]
            [Inspection task subtree, roundnode, style={draw, fill=blue!10}
            ]
        ]
        [Default behavior subtree, roundnode, style={draw, fill=blue!10}]
    ]
    \end{forest}
\end{adjustbox}
    \caption{The behavior tree structure used to switch between inspection tasks and a default behavior. The red nodes are later expanded as sub-trees based on the generated, back-chained, task specific behavior trees.}
    \label{fig:Main_Behavior_Tree}
\end{figure}
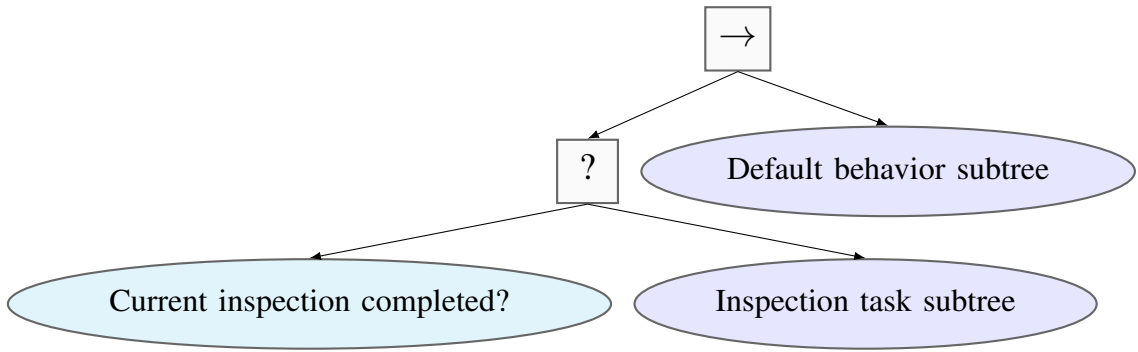

To enable the use of the back-chaining principle to switch between generated task-specific behavior trees, the behavior tree structure in Fig. \ref{fig:Main_Behavior_Tree} is used, specifically for performing inspection tasks with a default (when not allocated to a task) behavior of returning to a home location. 
Here, the behavior tree is structured considering only one task-specific sub-tree but could, without loss of generality, be extended to include more tasks by adding additional condition nodes to check which task is currently active.
\subsubsection{Task Specific Behavior Trees}
\begin{table*}[]
    \centering
    \caption{The library \(\mathcal{L}\) of actions and conditions used in the evaluation scenario to generate the task-specific behavior trees.}
    \begin{tabular}{|l|c|l|}
        \hline
        Action, \(\mathcal{A}_i\) & Precondition, \(\mathcal{A^{\mathrm{Pre}}}_{i, j}\) & Postcondition, \(\mathcal{A^{\mathrm{Post}}}_{i, k}\)\\
        \hline
        Set home location & - & Has home location \\
        Arm & - & Is armed \\
        Set flight mode OFFBOARD & - & In OFFBOARD mode \\
        Takeoff & Has home location, Is armed, In OFFBOARD mode & Is Flying \\
        Follow path & Is flying & At goal point\\
        Update path & - & Has path \\
        \hline
    \end{tabular}
    \label{tab:action_library}
\end{table*}

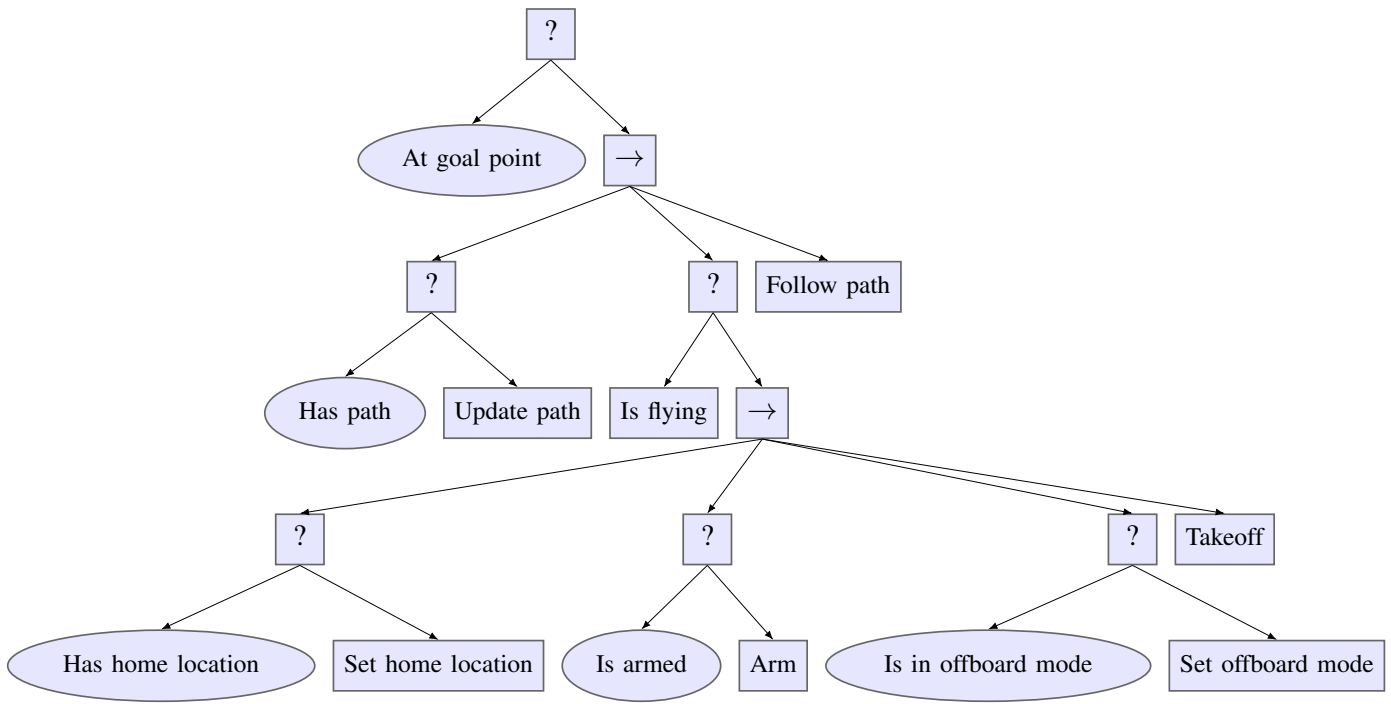
\begin{figure}
    \centering
    \begin{adjustbox}{max width = \columnwidth}
    \begin{forest}
    for tree={align=center, edge={-{Latex[length=1.5mm]}, draw}, parent anchor=south,child anchor=north, l=60pt}
    [\large ?, rectanglenode, style={draw, fill=blue!10}
        [At goal point, roundnode, style={draw, fill=blue!10}]
        [\large \(\rightarrow\), rectanglenode, style={draw, fill=blue!10}
            [\large ?, rectanglenode, style={draw, fill=blue!10}
                [Has path, roundnode, style={draw, fill=blue!10}]
                [Update path, rectanglenode, style={draw, fill=blue!10}]
            ]
            [\large ?, rectanglenode, style={draw, fill=blue!10}
                [Is flying, rectanglenode, style={draw, fill=blue!10}]
                [\large \(\rightarrow\), rectanglenode, style={draw, fill=blue!10}
                    [\large ?, rectanglenode, style={draw, fill=blue!10}
                        [Has home location, roundnode, style={draw, fill=blue!10}]
                        [Set home location, rectanglenode, style={draw, fill=blue!10}]
                    ]
                    [\large ?, rectanglenode, style={draw, fill=blue!10}
                        [Is armed, roundnode, style={draw, fill=blue!10}]
                        [Arm, rectanglenode, style={draw, fill=blue!10}]
                    ]
                    [\large ?, rectanglenode, style={draw, fill=blue!10}
                        [Is in offboard mode, roundnode, style={draw, fill=blue!10}]
                        [Set offboard mode, rectanglenode, style={draw, fill=blue!10}]
                    ]
                    [Takeoff, rectanglenode, style={draw, fill=blue!10}]
                ]
            ]
            [Follow path, rectanglenode, style={draw, fill=blue!10}]
        ]
    ]
    \end{forest}
\end{adjustbox}
    \caption{The behavior tree generated through the back-chaining approach, for executing  inspection tasks. The Algorithm \ref{alg:generate_tree} was used to generate the behavior tree, with `At goal point' as the required condition.}
    \label{fig:bt_inspection_task}
\end{figure}

\begin{figure}
    \centering
    \begin{adjustbox}{max width = \columnwidth}
    \begin{forest}
    for tree={align=center, edge={-{Latex[length=1.5mm]}, draw}, parent anchor=south,child anchor=north}
    [\large ?, rectanglenode, style={draw, fill=blue!10}
        [Landed?, roundnode, style={draw, fill=blue!10}]
        [\large \(\rightarrow\), rectanglenode, style={draw, fill=blue!10}
            [Hold position, rectanglenode, style={draw, fill=blue!10}]
            [Fly to home location, rectanglenode, style={draw, fill=blue!10}]
            [Land, rectanglenode, style={draw, fill=blue!10}]
        ]
    ]
    \end{forest}
\end{adjustbox}
    \caption{
    Generated behavior tree for executing the default behavior. It is executed to allow for the agent, if no task is allocated for a specified time, to return to its home location to land and wait for being allocated to new tasks.}
    \label{fig:bt_default_behavior}
\end{figure}
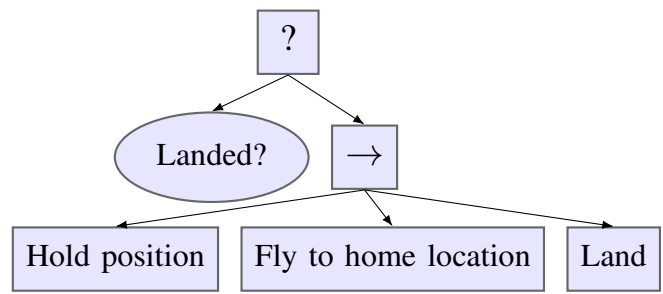

The library of available actions \(\mathcal{L}\) utilized to generated the required task specific behavior trees, as detailed in Algorithms \ref{alg:generate_tree} and \ref{alg:exapnd_condition}, can be seen in Table \ref{tab:action_library}. The resulting task specific behavior trees utilized in the inspection mission, one for executing a task and one for defining the agents default behavior, are shown in Fig. \ref{fig:bt_inspection_task} and Fig. \ref{fig:bt_default_behavior}. 

\begin{figure}[ht]
    \centering
    \begin{subfigure}{0.465\columnwidth}
        \centering
        \includegraphics[width = \textwidth]{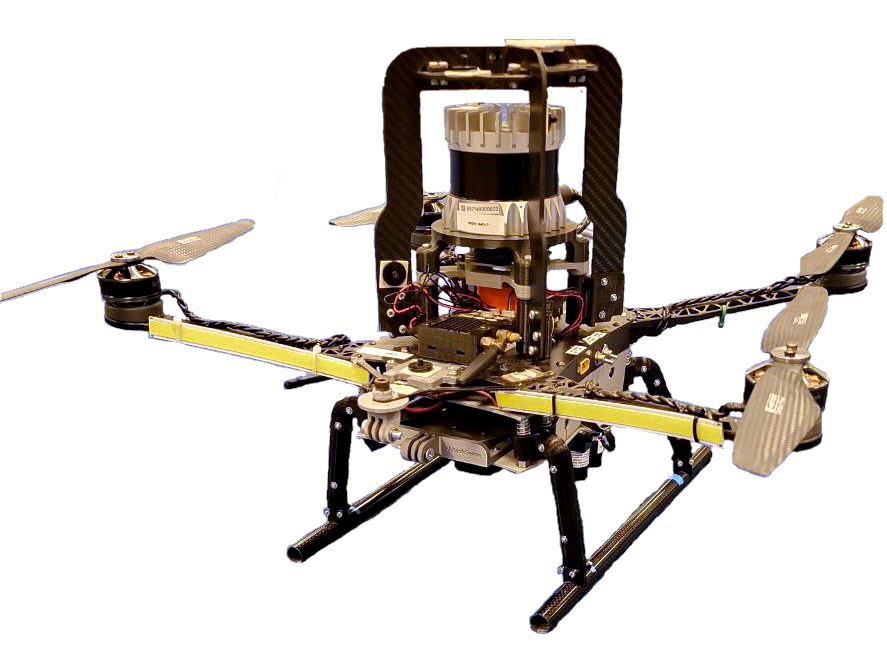}
        \caption{}
    \end{subfigure}%
    \hfill
    \begin{subfigure}{0.535\columnwidth}
        \centering
        \includegraphics[width = \textwidth]{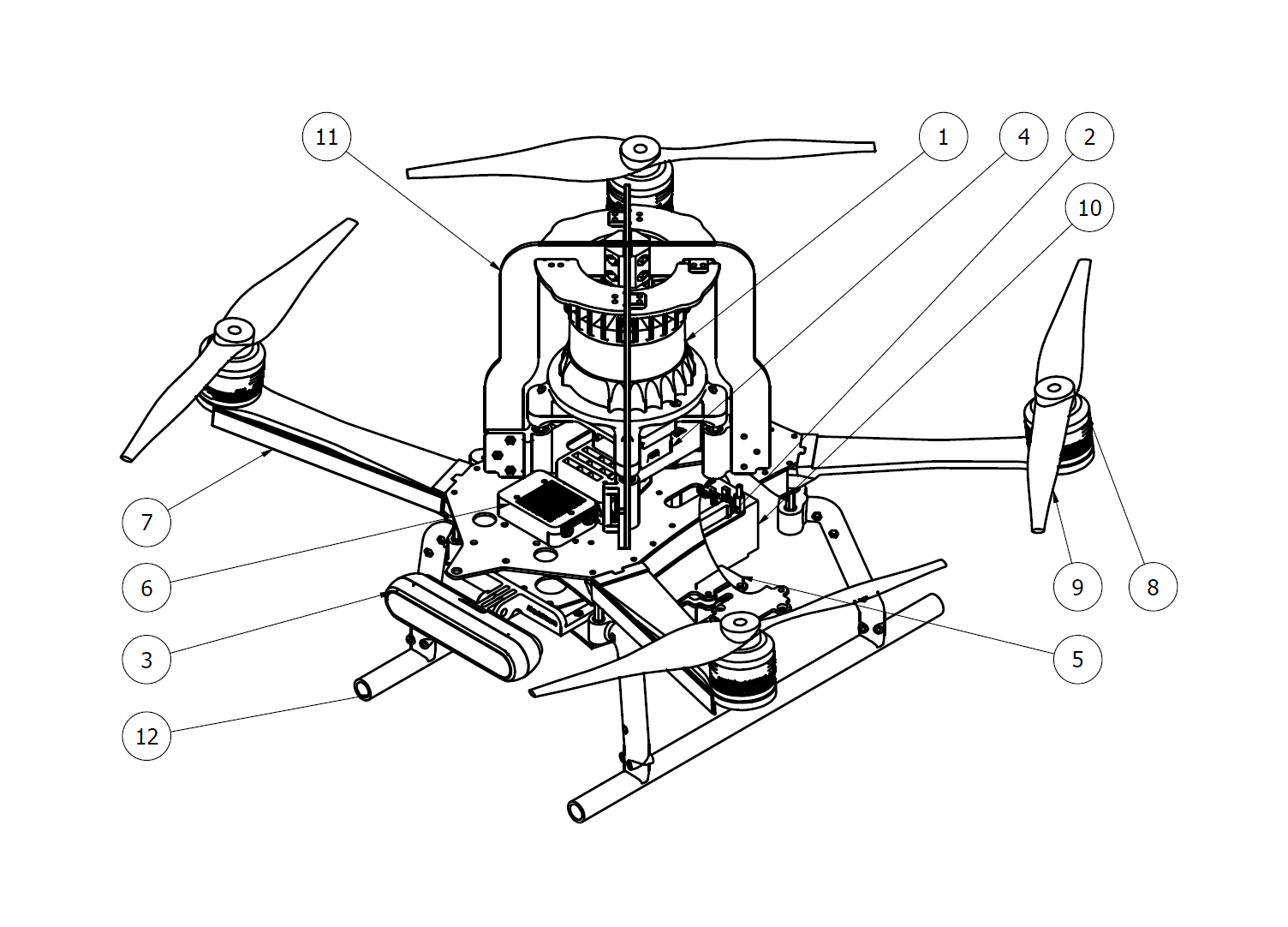}
        \caption{}
    \end{subfigure}%
    \hfill

    \caption{The platform used during the field deployment. In the main mission, three of the shown aerial vehicles were deployed to collaboratively execute the multi-agent mission. (a) shows an image of the real platform and (b) shows a schematic drawing with the following key parts specified: 1- Ouster OS1 3D LiDAR, 2- Intel NUC, 3- Intel Realsense D455, 4- Pixhawk Cube Flight Controller, 5- Garmin singlebeam LiDAR, 6- Telemetry module, 7- LED strips, 8- T-motor MN3508 kV700, 9- 12.5 in Propellers, 10- Battery, 11- Roll cage, 12- Landing gear.}
    \label{fig:hardware_platform}
\end{figure}

%
%

\begin{figure*}[t]
    \centering
    \begin{subfigure}[t]{0.47\textwidth}
        \centering
        \includegraphics[width = \textwidth]{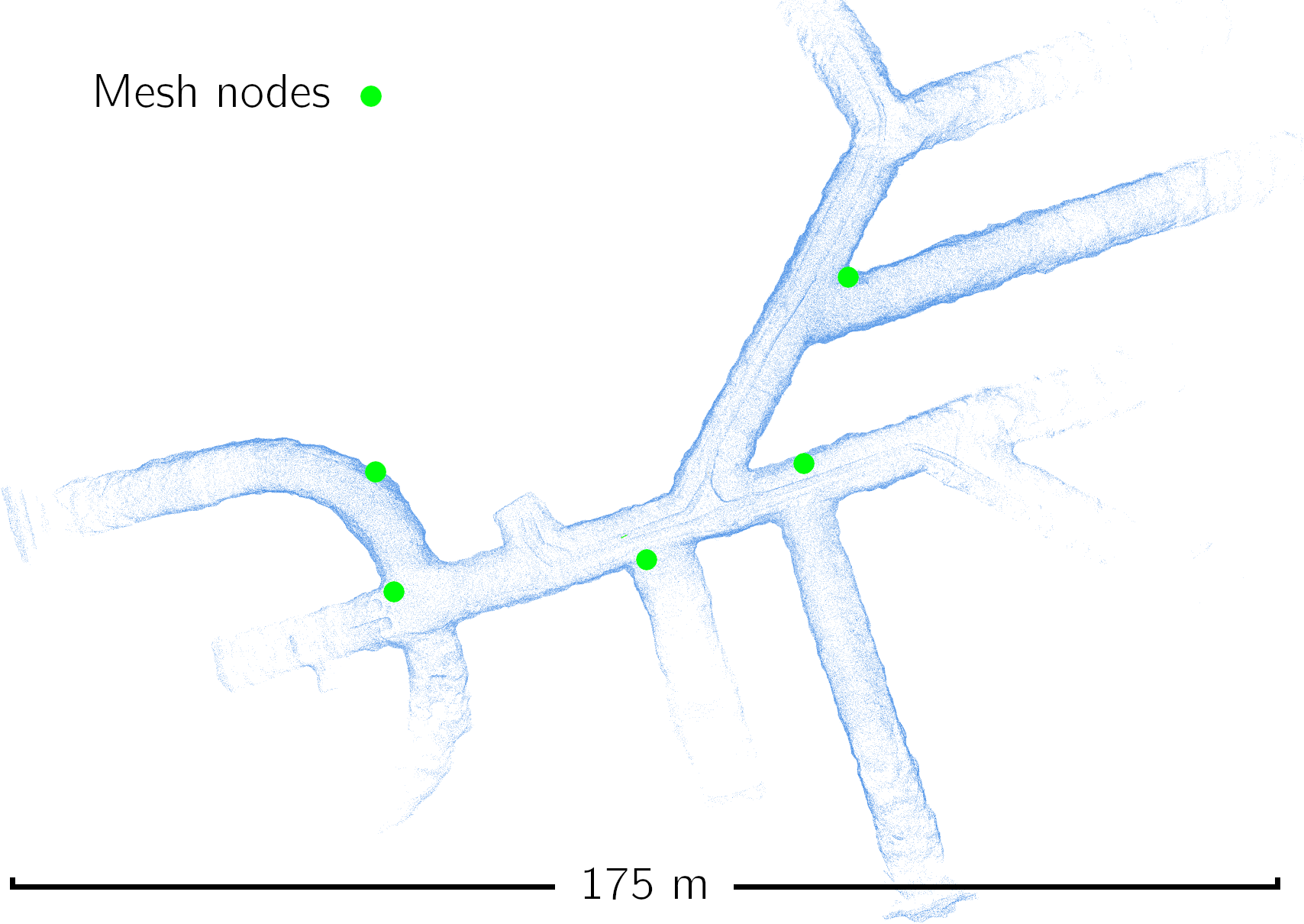}
        \caption{}
        \label{fig:mission_results_overview:a}
    \end{subfigure}%
    \hfill
    \begin{subfigure}[t]{0.47\textwidth}
        \centering
        \includegraphics[width = \textwidth]{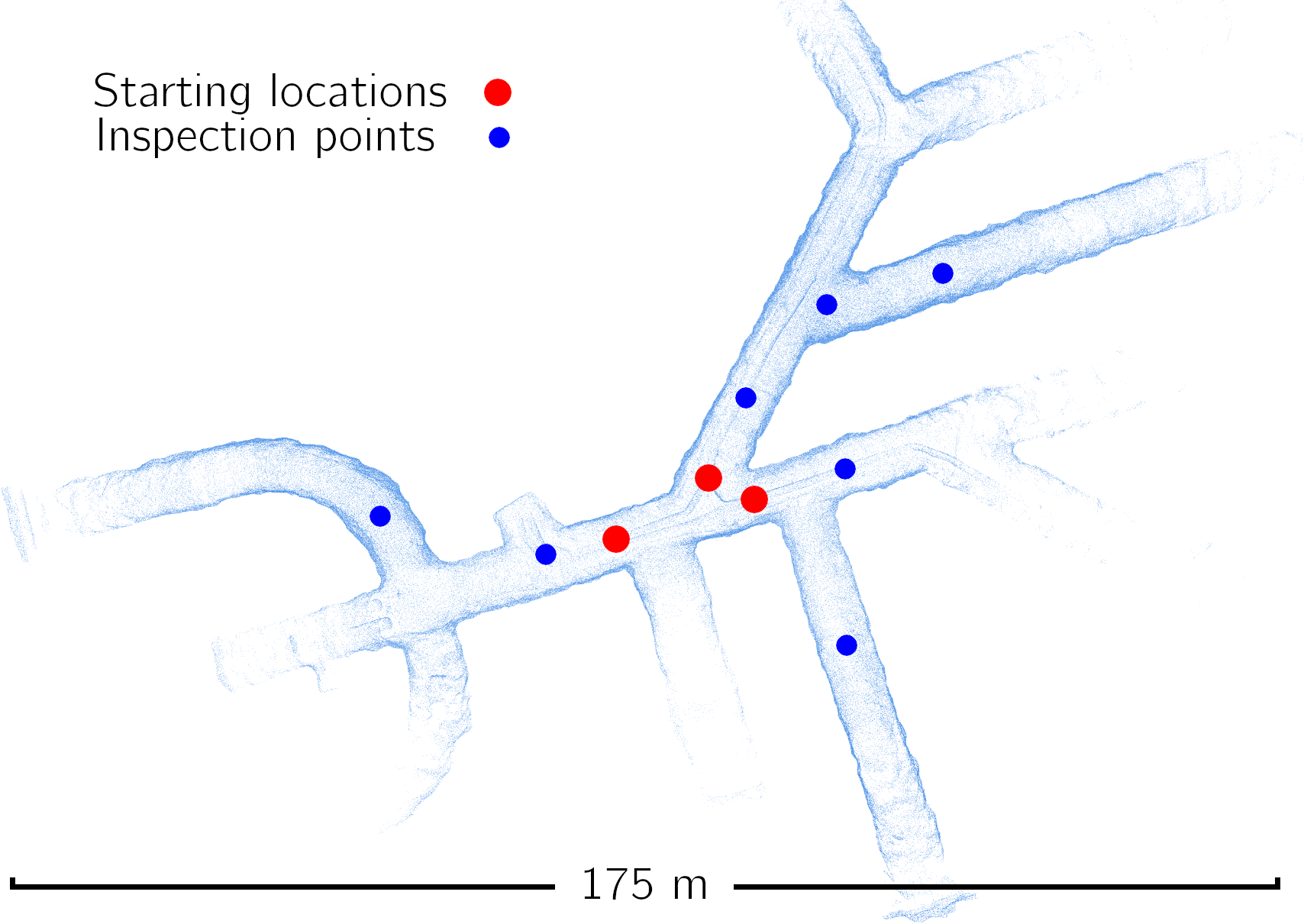}
        \caption{}
    \end{subfigure}%
    \caption{Data showing the scale of the underground environment where the multi-agent deployment was executed. (a) Shows the map and the locations of the deployed communication nodes and (b) illustrates the starting position and the locations for the inspection tasks that where executed during the mission. The map is the same as was used for localization and path planning during the mission.}
    \label{fig:mission_results_overview}
\end{figure*}

\begin{figure}[htb!]
    \centering
    \includegraphics[width=0.95\columnwidth]{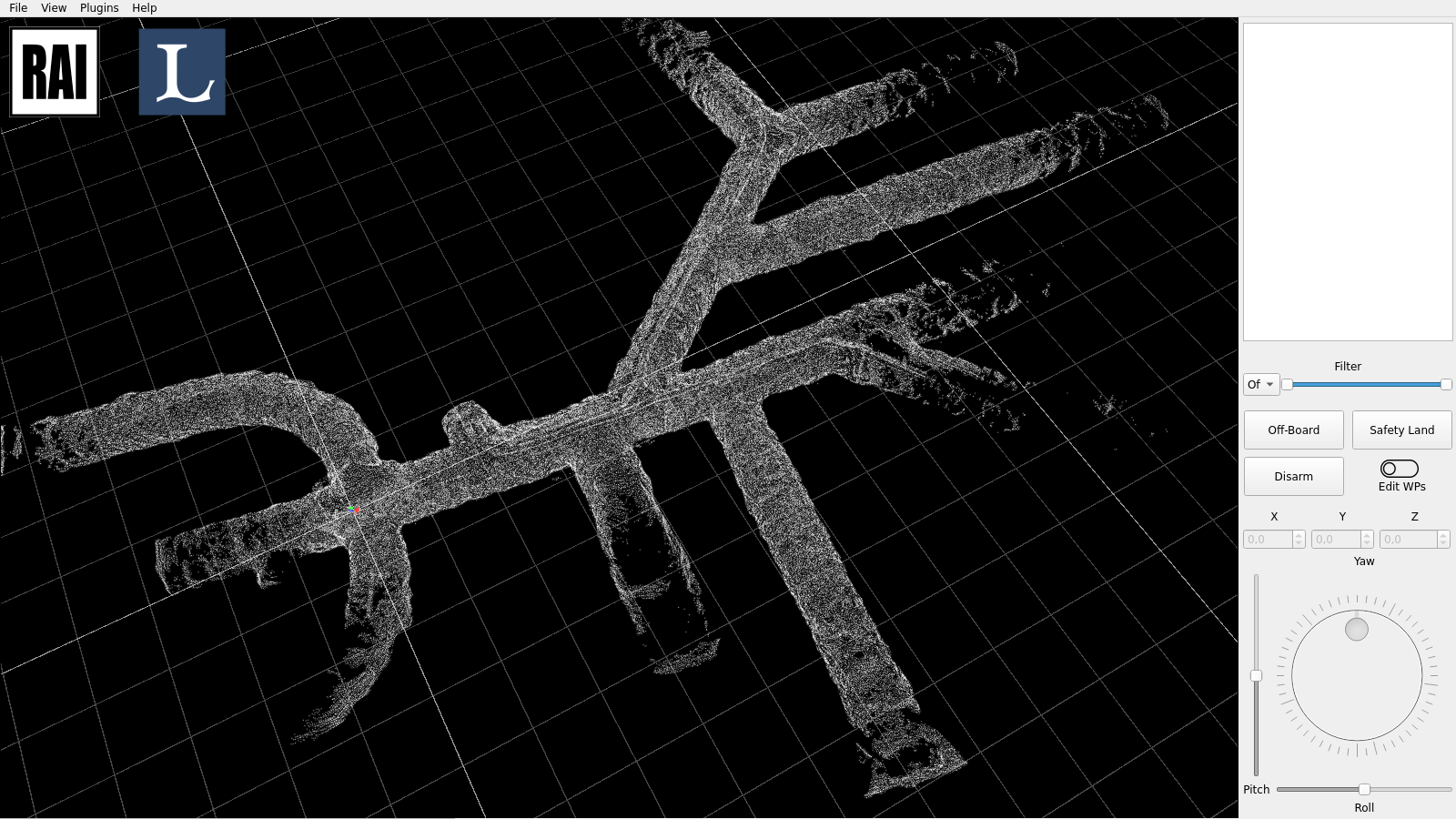}
    \caption{A view of the operators graphical interface where inspection tasks where added to initiate the mission and during the mission.}
    \label{fig:operator_ui}
\end{figure}
%
%

%
%
\section{Mission Preparations and Description}
\subsection{Hardware} \label{sec:hardware}
The three aerial platforms used for field demonstrations are custom-built drones designed by Robotics \& AI Group at Luleå University of Technology. The design and a schematic figure of one of the platforms is shown in Fig. \ref{fig:hardware_platform}. Most of the autonomy components of the proposed framework partly rely on information from the 3D LiDAR sensor since the drone is required to navigate and localize in completely dark environments. In the development process, we evaluated multiple LiDAR systems and settled on the Ouster OS1 32-beam, due to its robustness when operating in dusty environments commonly encountered in underground mines.

The UAV also carries a Pixhawk Cube FCU (Flight Control Unit) that handles all low level control, with its internal IMUs (Intertial Measurement Unit) also providing acceleration and roll/pitch data to the Lidar-Intertial localization framework. All computation is handled by an onboard Intel NUC. 
Not shown in the figure is a GoPro camera that was added to the front of the UAV to provide visual imagery during inspection missions. The camera was selected simply for its large internal storage, low weight and ease of use.

\subsection{Mission Set-up}\label{subsec:mission_set_up}
\subsubsection{Preparation Phase}

\begin{figure}[h]
    \centering
    \includegraphics[width=0.75\columnwidth]{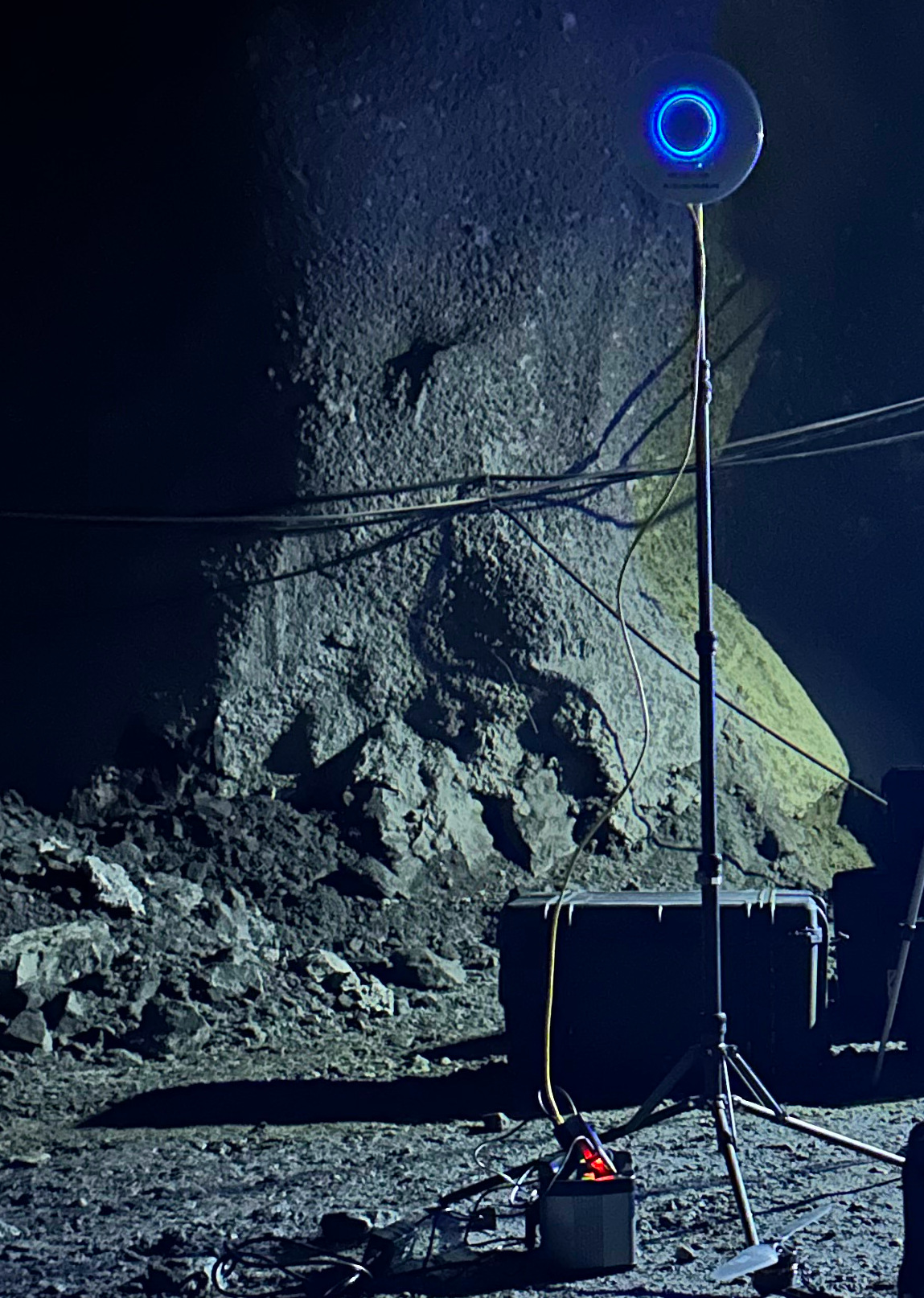}
    \caption{One of the nodes in the mobile Wi-Fi mesh used to provide communication for the task coordination. The Ubiquiti U6 Pro is mounted on a tripod and connected to a battery to allow for easy and quick deployment in the underground environment.}
    \label{fig:mesh_node}
\end{figure}

Before the mission begins, thorough preparation is essential, with respect to setting up of the UAV agents, communication infrastructure and deployment protocols. This involves ensuring that all agents are equipped with the required sensors, such as 3D LiDAR for localization and scanning and cameras for visual feedback and visual inspection, and communication equipment. A portable, battery powered Wi-Fi mesh, detailed in section \ref{sec:wifi-mesh}, is set up to establish communication between the agents and the auctioning system for reactive task allocation among the agents during the mission and to allow for adding new tasks on-the-fly. Fig. \ref{fig:mesh_node} shows an image of one of the five Wi-Fi nodes deployed in the subterranean environment. Each Wi-Fi node consists of: 1) the communication hardware, the Wi-Fi 6 enabled access point U6 Pro from Ubiquiti Inc. 2) a battery pack providing 48 V 
power over Ethernet and 3) a tripod raising the node to an efficient height and ensuring that the battery pack is kept away from the wet/muddy floor and good communication is established between the nodes.

The execution of large-scale robotics experiments requires a team effort and setting up of modes of coordination and communication protocols is important. A safety briefing is conducted for the entire team, detailing the mission objectives, communication protocols, and emergency procedures, to ensure that everyone is aware of their roles and responsibilities and how to react in case of an accident. 

\subsubsection{Deployment Phase}

The deployment phase begins with the setup of a launching area, which serves as the control center during the mission. This is typically established at a safe, stable location of the mine. From this area, the mission is started and the aerial agents are launched. Additionally, initial system checks are performed to assure that the mission can be launched.

To enable the use of multiple agents in a shared environment, a mapping session is performed prior to the inspection phase. The 3D map can be constructed either with a handheld device or by an autonomous agent~\cite{patel2022ref} through an exploration mission. Once the point cloud map is constructed and loaded, the agents are re-localized within it, in order to have a common coordinate frame, either manually by the operator or automatically as described in subsection~\ref{subsec:localization}.
Then, the desired inspection locations, including the type of inspections that need to be performed during this specific mission, are chosen and the inspection mission is ready to be carried out by the aerial agents. 

\subsubsection{Inspection Phase}
During the inspection phase, the aerial agents execute the assigned inspection tasks by navigating through the environment to the target locations. The path to the target locations is generated using the D$^*_+$ risk-aware path planner ((section \ref{subsec:Risk_Aware_Path_Planning})) and the paths are tracked using NMPC, with APF providing reactive local avoidance and an additional safety layer (section \ref{subsec:NMPC_APF}). 
As the inspection tasks are carried out, the agents transmit the gathered data back to the control center to be viewed in real time. This enables real-time mission adjustments, addition of new inspection tasks on-the-fly, and removal of inspection tasks that are no longer necessary. 

The mission concludes with a debriefing to review the mission's outcome, discuss any challenges faced and identify areas for improvement. By following this structured approach, safe and efficient inspection missions are assured and continuous improvements to future inspection missions can be achieved.
\subsection{Limited Infrastructure} \label{sec:wifi-mesh}
Each agent has its own autonomy stack on board, in the sense that they can operate without any external infrastructure. However, the auction-based task allocation system relies on continuous communication to be able to correctly add new tasks to the system and assign tasks to the correct agents during the mission execution. To cater to such communication needs, a local Wi-Fi mesh is utilized to ensure reliable communication between the agents and the base station.
Because the agents are in charge of their local autonomy, and minimally interact with the central auctioning system to send bids and receive task assignments, the entire system can be run over limited bandwidth. 
Data such as odometry and pointclouds are required by the operators of the multi-agent system, to visualize the mission execution and to meaningfully interact with the entire system by adding new tasks and follow/monitor the progress. The Wi-fi mesh is also used to stream such data over the network. 
\subsubsection{Mobile Reconfigurable Wi-Fi Mesh}
During the field experiments in a mine, a mobile mesh network was deployed to provide Wi-Fi connectivity between the agents, the auctioning system and the operator. 
The mesh network consists of five Wi-Fi repeaters (Fig. ~\ref{fig:mesh_node}) positioned as shown in Fig.~\ref{fig:mission_results_overview:a}. The nodes are placed so that each repeater is in the line-of-sight of the next one, to ensure high RSSI for a reliable and high capacity communication between the nodes. 

%
%
\section{Field Deployment (Experiments/Evaluation)}\label{sec:field_deployment}
This section describes the field environment and results from a field mission, where an aerial multi-agent system was deployed to demonstrate the capabilities of the proposed task allocation and task execution system in the field. The field deployment was executed with our mining industry collaborators at Luossavaara-Kiirunavaara AB (LKAB), through the Sustainable Underground Mining (SUM) Academy programme.



%
%
\subsection{Environment/Mission Description}
\begin{figure*}
    \centering
    \begin{subcaptionblock}[t]{0.49\textwidth}
        \centering
        \includegraphics[width = \textwidth]{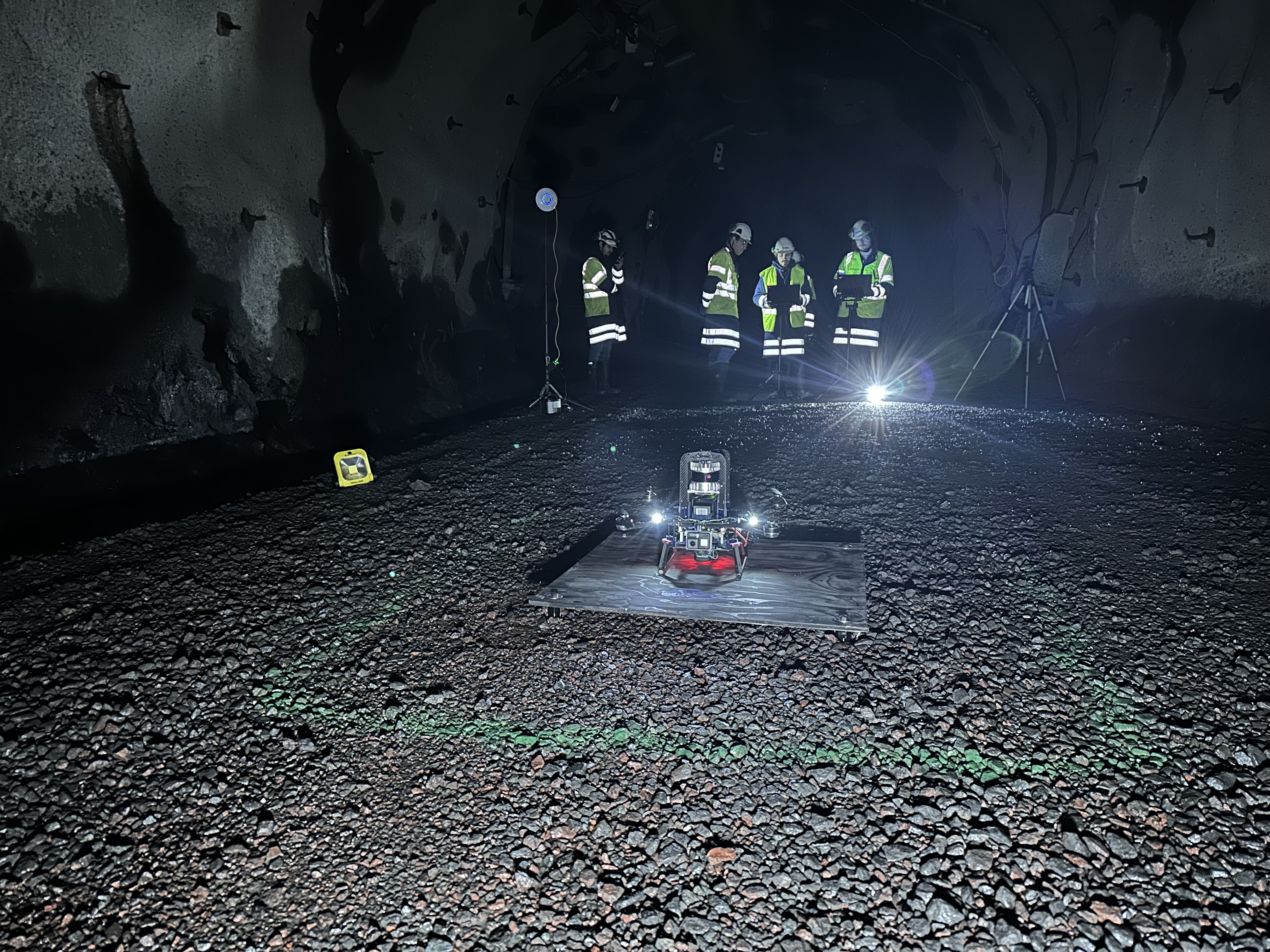}
        \caption{}
    \end{subcaptionblock}%
    \hfill
    \begin{subcaptionblock}[t]{0.49\textwidth}
        \centering
        \includegraphics[width = \textwidth]{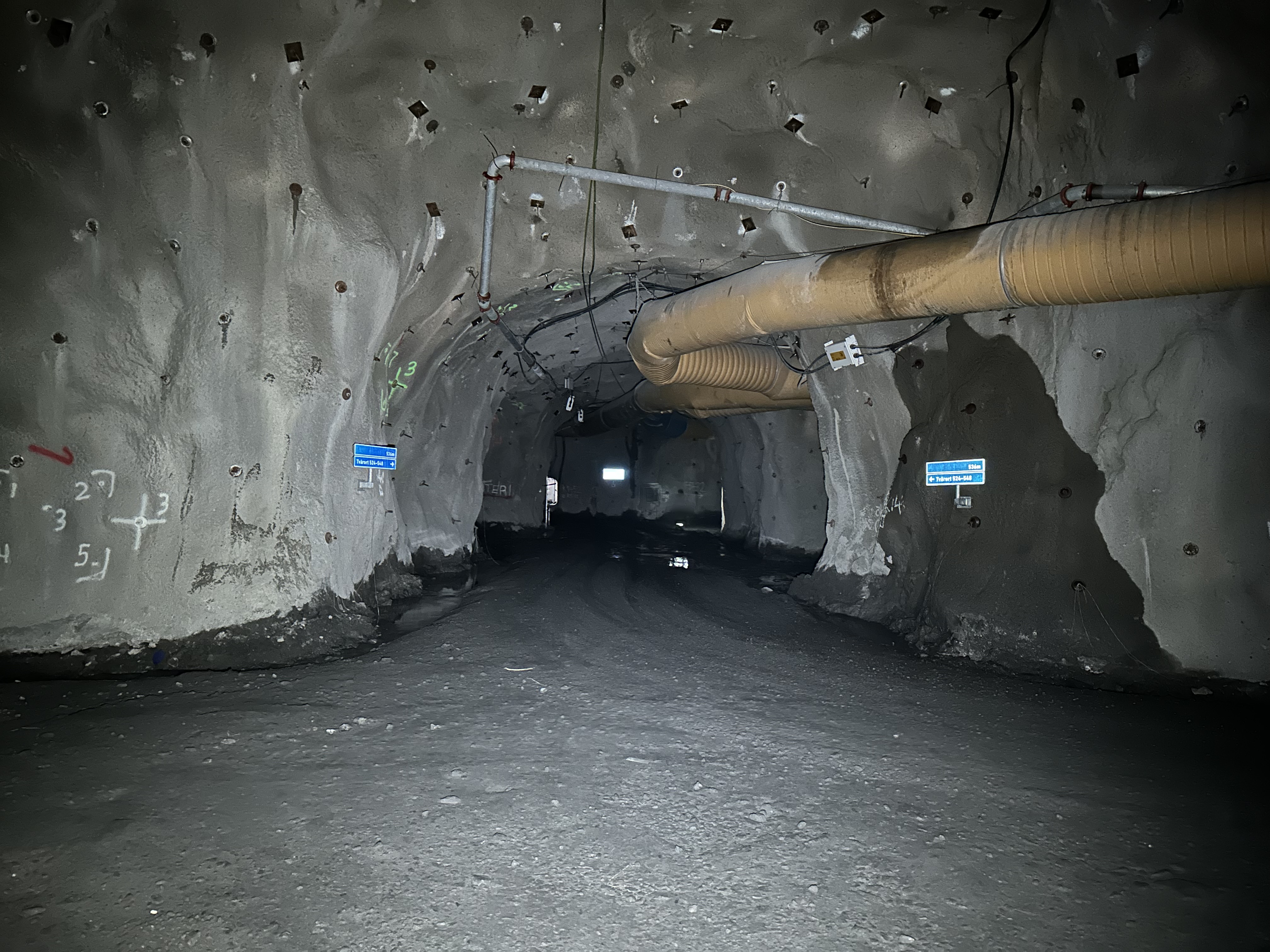}
        \caption{}
    \end{subcaptionblock}%
    \\
    \begin{subcaptionblock}[t]{0.49\textwidth}
        \centering
        \includegraphics[width = \textwidth]{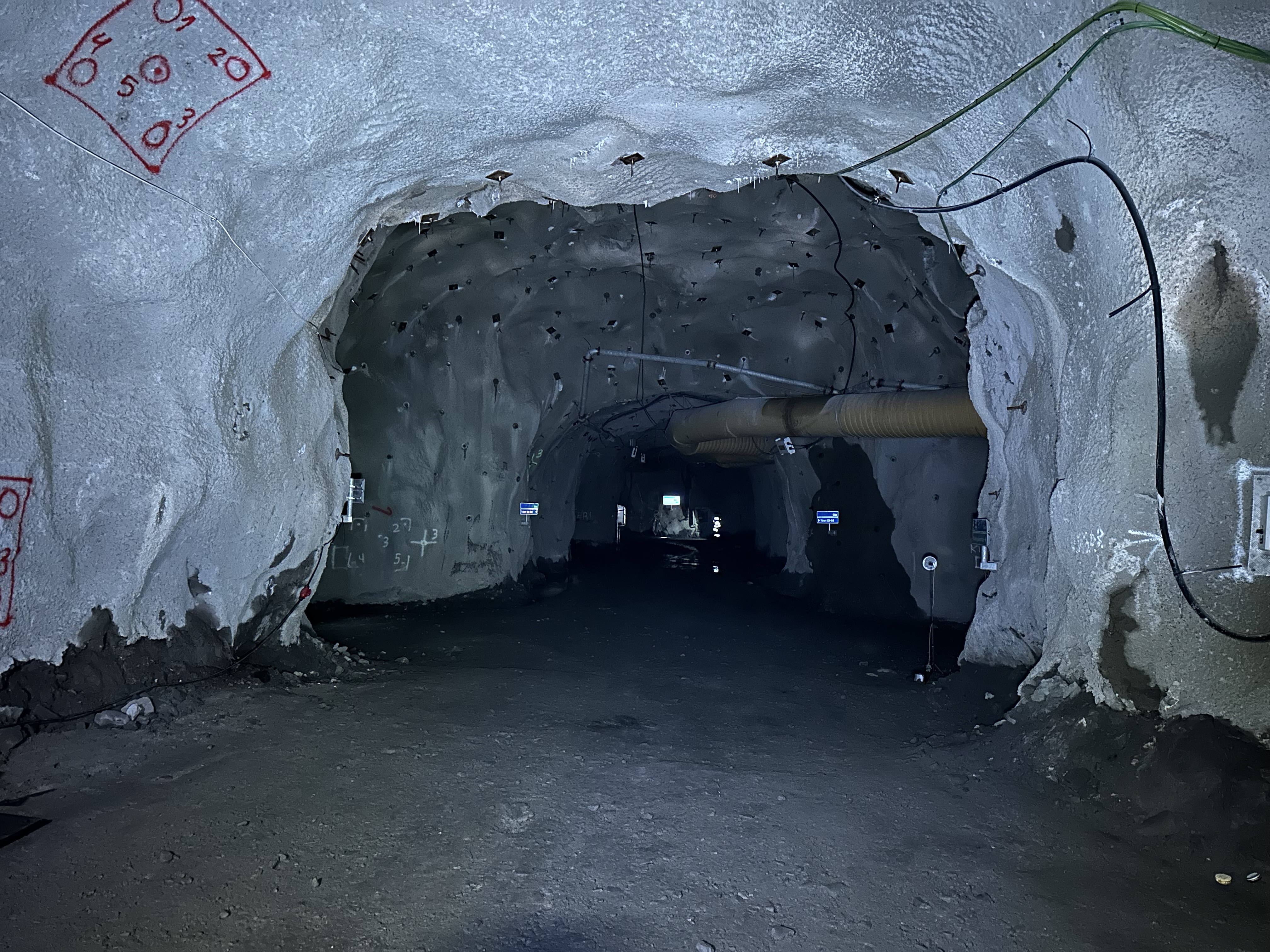}
        \caption{}
    \end{subcaptionblock}%
    \hfill
    \begin{subcaptionblock}[t]{0.49\textwidth}
        \centering
        \includegraphics[width = \textwidth]{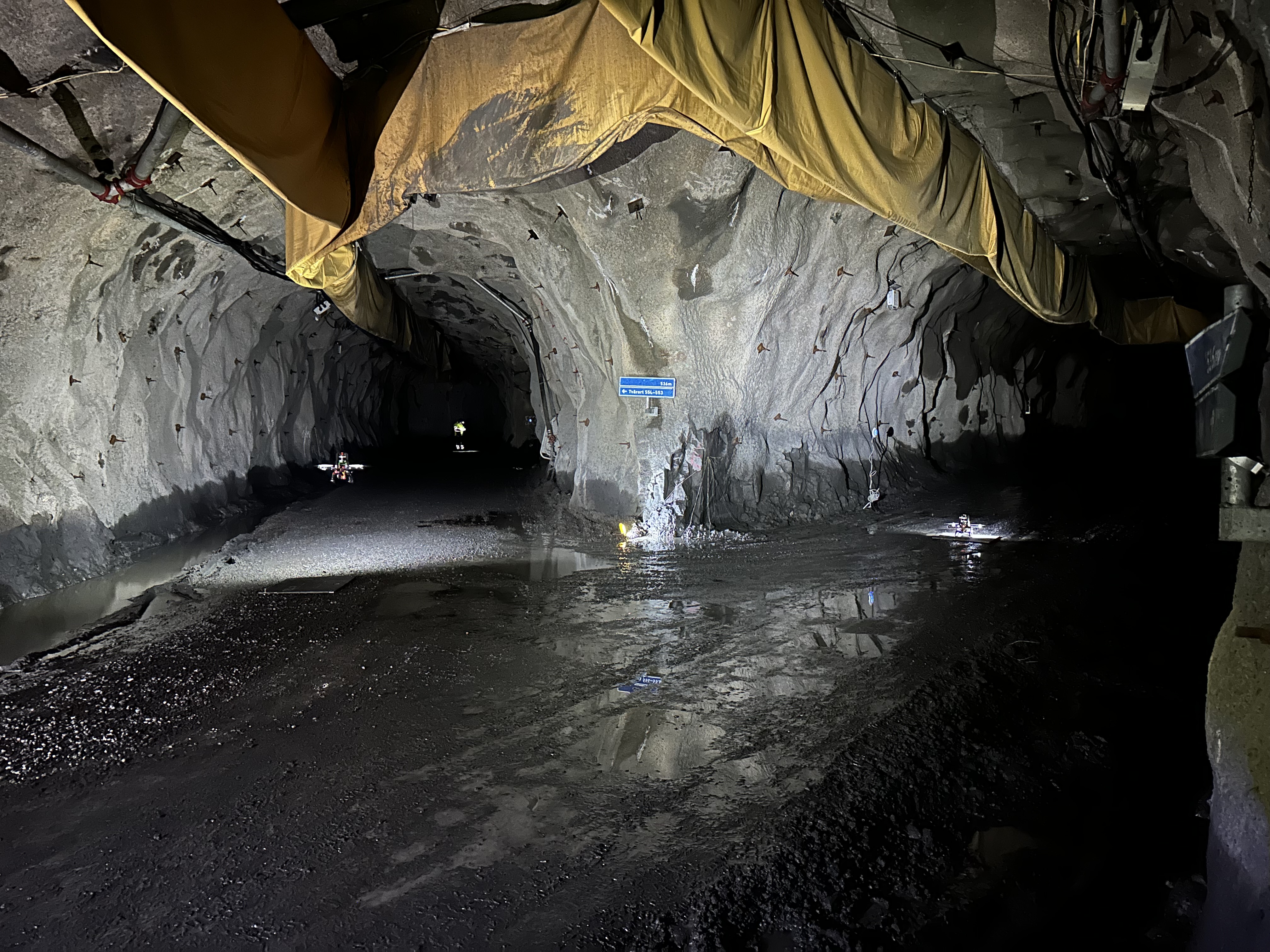}
        \caption{}
    \end{subcaptionblock}%
    \caption{Pictures showing the mining environment where the system was deployed and evaluated. (a) shows one of the aerial agents used to perform the inspection tasks prior to takeoff, (b) -- (d) shows three snapshots of the tunnels in the underground mine where the evaluation mission was performed.}
    \label{fig:mission_environment}
\end{figure*}

The proposed multi-agent architecture is evaluated in a mission where a fleet of aerial agents are deployed (detailed in section \ref{sec:hardware}), in the challenging conditions of an underground mine to perform routine inspections tasks, with the  inspection locations being set by a mine operator. These inspections are vital for identifying potential hazards and maintaining overall safety standards. The mission is meticulously organized into three key phases: preparation, deployment and inspection, described in section \ref{subsec:mission_set_up}. The main mission included three aerial agents performing inspection tasks in a collaborative fashion. An operator initiated the mission by specifying the inspection locations, and the operator had the authority to add new tasks during the execution, through the  interface shown in Fig. \ref{fig:operator_ui}. The environment where the multi-agent system was deployed is illustrated in Fig. \ref{fig:mission_environment} and comprises of typical subterranean mining environment with interconnected tunnels. 

\begin{table}[t]
    \centering
    \caption{Table summarizing important stats from the experiment. Total distance traveled by all agents, mission duration, number of locations inspected.}
    \label{tab:main_statistics}
    \begin{tabular}{|l|c|}
        \hline
        Parameter & Value \\
        \hline
        Mission duration & 311 s\\
        Total distance covered & 143.7 m \\
        Number of agents & 3 \\
        Number of inspection points & 7 \\
        \hline
    \end{tabular}
\end{table}

\begin{table}[t]
    \centering
    \caption{Table showcasing the execution of the inspection tasks. Includes the time when the tasks was added to the pool of available tasks, the time when a task was considered completed, the time it took from when an agent was allocated to a task until it was completed and what agent completed what task.
    }
    \label{tab:task_information}
    \begin{tabular}{|c|c|c|c|c|}
        \hline
        Task & Added [s] & Finished [s] & Execution Time [s] & Agent \\
        \hline
        \(T_1\) & 0 & 69 & 69 & \(R_1\)\\
        \(T_2\) & 0 & 46 & 46 & \(R_1\) \\
        \(T_3\) & 0 & 68 & 68 & \(R_2\) \\
        \(T_4\) & 0 & 115 & 115 & \(R_2\) \\
        \(T_5\) & 0 & 50 & 50 & \(R_3\) \\
        \(T_6\) & 0 & 103 & 103 & \(R_3\) \\
        \(T_7\) & 116 & 157 & 41 & \(R_2\) \\
        \hline
    \end{tabular}
\end{table}

%


%
%
\subsection{Field Deployment Results}
\begin{figure*}[t]
    \centering
    \includegraphics[width = 0.85\textwidth]{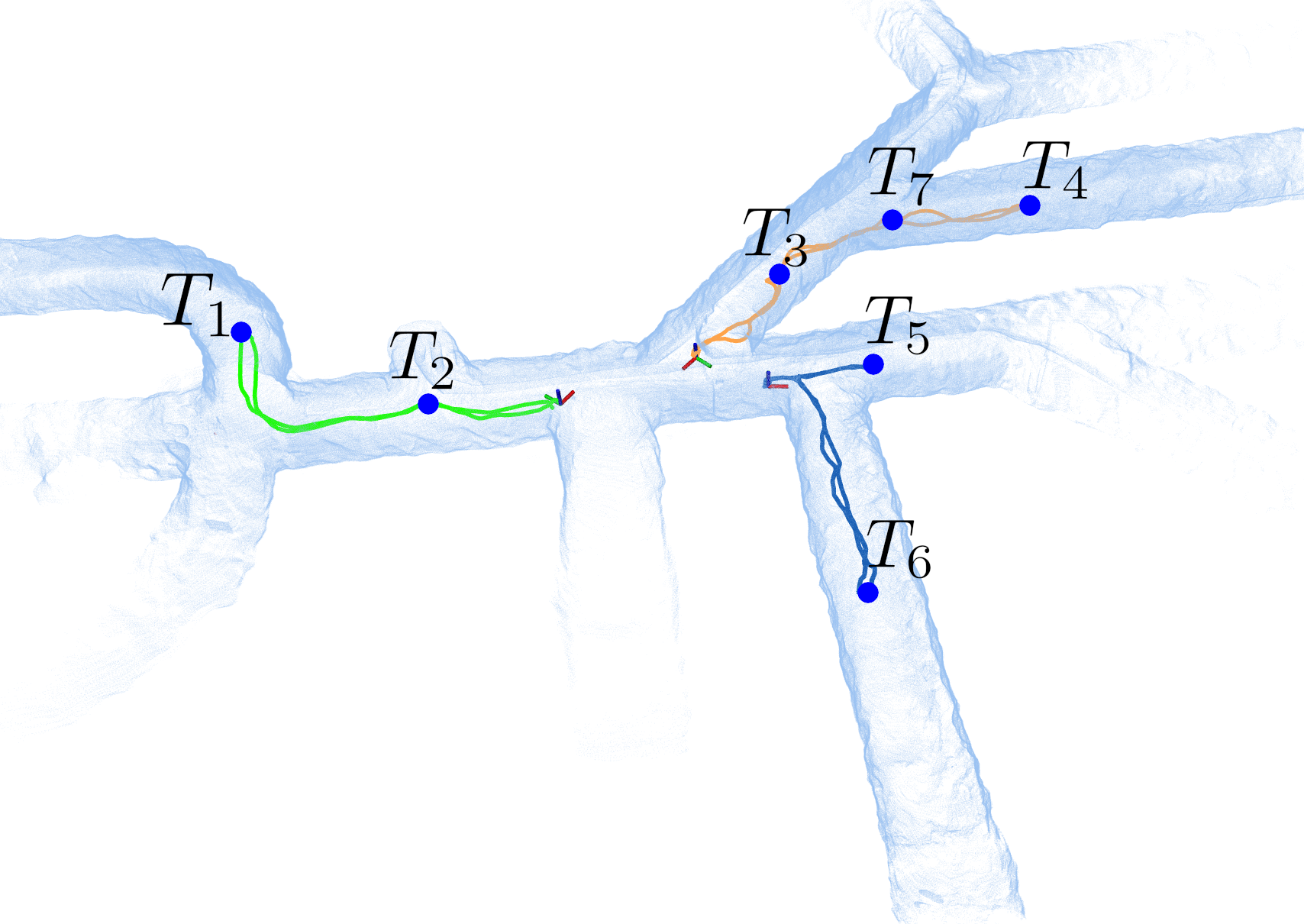}
    \caption{Illustration to show the results from the inspection mission, with the traversed paths and the inspection locations highlighted, from the operators viewpoint with the autonomous agents localized in a common frame of reference. In total, seven discrete inspection tasks, indicated in the figure, where completed by the three aerial agents during the mission.}
    \label{fig:mission_operators view}
\end{figure*}

In total, seven inspection tasks where executed by a team of three aerial agents. The key parameters of the mission is summarized in Table \ref{tab:main_statistics}.
The location of the seven inspection tasks \(\{T_1, \dots, T_7\}\) and the agents \(\{R_1, R_2, R_3\}\) are indicated in Fig. \ref{fig:mission_environment}. The times when tasks where added by the operator, the times when the task were completed and the execution information of all the inspection tasks are detailed in Table \ref{tab:task_information}. The aerial agents autonomously navigated a complex environment using on-board sensing and computation, while all the communication related to task coordination was enabled through the deployed Wi-Fi mesh. The field deployment of the multi-agent system in the underground mine successfully demonstrated the performance and usability of the framework. 

A video to show the system's performance can be found at the following link: \href{https://youtu.be/4eyRCCRAEYg?si=79lJcdv8x6Ju3jAp}{https://youtu.be/4eyRCCRAEYg?si=79lJcdv8x6Ju3jAp}. The video shows the harsh mining environment and the performance of the aerial agents during the execution of the inspection tasks. In Fig. \ref{fig:mission_operators view},  an overview of the entire mission is presented from the operators viewpoint, along with the agents' initial locations, the location of the inspection tasks and the paths traversed by the agents to reach the target inspection points. The deployment highlighted the flexibility and ease-of-use, confirming the potential for enhancing routine inspection in large-scale production industrial settings.

%
%
\subsection{Lessons Learned}

The deployment of a multi-agent system in a large-scale subterranean environment for inspection tasks has yielded multiple valuable lessons, both practical, such as dealing with perpetual hardware issues, and abstract, like designing the overarching software structure. Firstly, the importance of reliable communication became evident, as data loss can significantly impact coordination between drones. Secondly, the robustness of advanced navigation, localization, and path planning was thoroughly tested in the field. This work, building on previous efforts in path planning and navigation, allowed for quick deployments due to the high level of robustness already achieved in those areas. From a theoretical perspective, structuring the agents' available actions and then synthesizing a behavior tree for task execution proved efficient for both developing and improving certain parts. Additionally, it provided a standardized way to monitor individual agents during task execution. Similar lessons were learned from the design of task allocation, where a centralized task coordinator and independent task execution proved to be a good compromise in an environment where some communication infrastructure was available. This approach was taken, due to the specific scenario considered, to be able to provide a near real-time stream of data back to the operator.

One important goal was to handle the dynamic scenario where an operator could add new tasks into the system at any point. A centralized approach was chosen, which proved beneficial as it provided an intuitive way to foresee system behavior and allowed us to achieve our objectives. However, some limitations were encountered, especially related to reactive planning with a very short horizon. Occasionally, inefficient solutions emerged over time, highlighting the need for a real-time and computationally efficient way to incorporate either task clustering or planning into the task coordination architecture to address this issue.

\section{Conclusion}\label{sec:conclusion}


The proposed multi-agent framework for deploying aerial agents in an underground mine represents a significant advancement in industrial automation and autonomous inspection capabilities. By enabling full autonomy, this system allows for efficient, precise, and reliable inspection of complex and dangerous environments, greatly reducing the risk for human operators and enhancing the overall safety of mining operations. Several key capabilities, including autonomous inspection, flexible task allocation and adaptability has been presented during the presented large-scale mission. Furthermore, the framework demonstrated a flexibility to accommodate high-level inspection tasks, both before and during execution, by a mining operator specifying positions on an easy-to-use interface. This ensures that the system could meet the industrial needs and be used in production scenarios.

In summary, the autonomous multi-agent framework enhances current inspection capabilities in subterranean environments and also sets the stage for broader applications in industrial automation due to the ability of the framework to generalize the type of both tasks and robotic platforms.

\section{Acknowledgements}\label{sec:acknowledgements}
The authors would like to acknowledge and thank their mining industry collaborators at Luossavaara-Kiirunavaara AB (LKAB). The field demonstrations included in this manuscript were enabled through the Sustainable Underground Mining (SUM) Academy programme, which is funded by LKAB and the Swedish Energy Agency, and were executed at a depth of over 500m in an active production area of the mine located in Kiruna, Sweden.


 
%

\bibliographystyle{./bibtex/bib/IEEEtran}
\bibliography{./bibtex/bib/IEEEabrv, mybibfile}

\newpage

 




\vfill

\end{document}